\documentclass[conference]{IEEEtran}
\IEEEoverridecommandlockouts

\usepackage{url}
\usepackage{amsthm}
\usepackage{afterpage}
\usepackage{cite}
\usepackage{soul}
\usepackage[normalem]{ulem}
\useunder{\uline}{\ul}{}
\usepackage{graphicx}
\usepackage{textcomp}
\usepackage{xcolor}
\usepackage{balance} 
\usepackage{bbm}
\usepackage{amsmath,amssymb,amsfonts}
\usepackage{epstopdf}
\usepackage{array}
\usepackage{booktabs}
\usepackage{color}
\usepackage{xcolor}
\usepackage{colortbl}
\usepackage{xspace}
\usepackage{multirow}
\usepackage{pifont}
\usepackage{enumitem}
\usepackage{bbding}
\usepackage{hyperref}
\usepackage{makecell}
\usepackage{microtype}
\usepackage{algorithm}
\usepackage{algorithmic}
\usepackage{fontawesome}
\usepackage{caption,setspace}
\usepackage[utf8]{inputenc}
\newcommand{\ssymbol}[1]{^{\@fnsymbol{#1}}}

\newtheorem{proposition}{Proposition}
\newtheorem{question}{Question}

\begin{document}

\title{Breaking the Entanglement of Homophily and Heterophily in Semi-supervised Node Classification}

\makeatletter

\author{
    \IEEEauthorblockN{
    Henan Sun$^\dagger$$^\ddagger$, 
    Xunkai Li$^\dagger$$^\ddagger$, 
    Zhengyu Wu$^\dagger$,
    Daohan Su$^\dagger$,
    Rong-Hua Li$^\dagger$, 
    Guoren Wang$^\dagger$}
    \IEEEauthorblockA{
    $^\dagger$ Beijing Institute of Technology, Beijing, China}
    \IEEEauthorblockA{
    magneto0617@foxmail.com,
    cs.xunkai.li@gmail.com,\\
    Jeremywzy96@outlook.com,
    dhsu@bit.edu.cn,
    lironghuabit@126.com,
    wanggrbit@gmail.com
    }
}

\maketitle

\footnotetext{$\ddagger$: These authors contributed equally to this work.}

\begin{abstract}
    Recently, graph neural networks (GNNs) have shown prominent performance in semi-supervised node classification by leveraging knowledge from the graph database. 
    However, most existing GNNs follow the homophily assumption, where connected nodes are more likely to exhibit similar feature distributions and the same labels, and such an assumption has proven to be vulnerable in a growing number of practical applications. 
    As a supplement, heterophily reflects dissimilarity in connected nodes, which has gained significant attention in graph learning. 
    To this end, data engineers aim to develop a powerful GNN model that can ensure performance under both homophily and heterophily. 
    Despite numerous attempts, most existing GNNs struggle to achieve optimal node representations due to the constraints of undirected graphs. The neglect of directed edges results in sub-optimal graph representations, thereby hindering the capacity of GNNs.
    To address this issue, we introduce AMUD, which quantifies the relationship between node profiles and topology from a statistical perspective, offering valuable insights for \underline{A}daptively \underline{M}odeling the natural directed graphs as the \underline{U}ndirected or \underline{D}irected graph to maximize the benefits from subsequent graph learning.
    Furthermore, we propose \underline{A}daptive \underline{D}irected \underline{P}attern \underline{A}ggregation (ADPA) as a new directed graph learning paradigm for AMUD.
    Empirical studies have demonstrated that AMUD guides efficient graph learning. 
    Meanwhile, extensive experiments on 16 benchmark datasets substantiate the impressive performance of ADPA, outperforming baselines by significant margins of 3.96\%.

\end{abstract}


\section{Introduction}
\label{sec: introduction}
     {Graph neural networks (GNNs) have garnered considerable attention within the data engineering community~\cite{guan2023homo_measure_icde2,li2023irreversible_graph_vldb2,he2023scaling_sigmod2}. 
     The prevalent usage of structured data in wide-ranging applications, such as recommendation~\cite{10.1007/s00521-022-07735-y_EHGCN,xia2023app_gnn_rec1, yang2023app_gnn_rec2}, bioinformatics~\cite{bang2023app_gnn_bio1, qu2023app_gnn_bio2, gao2023app_gnn_bio3}, and anomaly detection~\cite{tang2022app_detection1, chen2022app_detection2, duan2023app_detection3} attests to its importance.
     Furthermore, GNNs have achieved state-of-the-art performance in the semi-supervised node classification paradigm~\cite{wang2020gcnlpa,chen2020gcnii, gamlp}, credited to their intuitive exploitation of the knowledge from node profiles and topology stored in databases~\cite{ye2023homo_measure_icde1,wu2023billionambiguous_graph_de_vldb1,liu2023efficientintro_sigmod1}.}

     {Reflecting on the evolution of GNNs, in their early stages, researchers rely on the homophily assumption~\cite{guan2023homo_measure_icde3,yin2023surel+irreversible_graph_vldb3,li2023efficient_sigmod3}, where connected nodes are more likely to possess similar features and the same labels. 
     This empirical assumption, commonly observed in real-world applications~\cite{mcpherson200homophily_theory1,M2003Mixing_homophily_theory2,0Networks_homophily_theory3}, guides principles for designing GNNs. 
     During this period, the well-known message-passing emerged~\cite{wu2020gnn_survey1,zhou2022gnn_survey2,bessadok2022gnn_survey3}, giving rise to numerous simple yet effective methods that continue to be widely applied.

     However, as GNNs become increasingly deployed in the intricate applications~\cite{ma2021hete_gnn_survey1,luan2022hete_gnn_survey2,zheng2022hete_gnn_survey3} and researchers demand more powerful representations (i.e. for higher accuracy and robustness), heterophily, the opposite of homophily, has gradually come into focus.
     To this end, data engineers have explored optimal node representations from both the spectral~\cite{he2021bernnet,pmlr2022Jacobigcn, guo2023optimal_poly_gnn} and spatial perspectives~\cite{2022glognn,lee2023aerognn,yoo2023slimg} with the aim of developing a GNN capable of delivering strong performance under both homophily and heterophily. 
     Despite their effectiveness, it's essential to note that all the aforementioned GNNs have been designed exclusively for undirected scenarios. 
     Therefore, the following limitations hinder the progress of graph learning.

     \textbf{L1: Irreversible loss of graph representation.}
     From the perspective of graph representation, the neglect of directed edges results in information loss within the natural graphs, which constrains the capacity of GNNs to capture and express relational information for predictions. 
     In contrast, directed graphs (digraphs) are better suited for modeling complex topology, given their ability to capture intricate relationships between nodes.
     We notice that recent Dir-GNN~\cite{dirgnn_rossi_2023} and A2DUG~\cite{maekawa2023a2dug} have acknowledged this limitation and attempted to improve performance through directed modeling.
     Unfortunately, they lack a comprehensive discussion of topological characteristics (i.e., homophily and heterophily) and directed information.
     In other words, there is room for improvement in their model architectures that lack the capture of directed topology.

     \textbf{L2: Ambiguous graph-based data engineering.}
     In our investigation of existing undirected/directed GNNs, we found that their approach to datasets is ambiguous. 
     Specifically, since real-world graphs naturally exhibit directed edges, undirected GNNs coarsely transform directed edges into undirected ones to guarantee predictive performance.
     Moreover, some directed GNNs might perform undirected transformations on digraphs for edge-wise data augmentation.
     In this context, we have the following observations:
     (\textbf{O1}) For some datasets, undirected GNNs based on coarse undirected transformations outperform directed GNNs based on natural digraphs. 
     But for others, the situation is quite the opposite.
     (\textbf{O2}) For directed GNNs, the effectiveness of undirected edge-wise data augmentation is uncertain.
     We attribute this to the entanglement of homophily and heterophily concealed beneath directed edges, which hinders graph data engineering within the narrow scope of undirected scenarios, highlighting the need for further investigating the inherent connection between nodes and topology.

    {To address the above limitations, we propose AMUD, which employs a statistical perspective to quantify the correlation between node profiles and topology. 
    It determines whether to retain the inherent directed edges in the original data, with the aim of maximizing the benefits for the subsequent training.
    After that, we suggest utilizing the undirected and directed GNNs to handle the undirected and directed output of AMUD, respectively, as methods specifically designed for undirected graphs exhibit performance advantages compared to directed GNNs.
    In comparison, our proposed ADPA discovers suitable directed patterns (DPs) and utilizes two hierarchical attention mechanisms to achieve multi-scale message aggregation, which applies to the both undirected and directed output of AMUD.}

    \textbf{Our contributions.}
    (1) \textit{\underline{New Perspective.}} 
    To the best of our knowledge, this paper is the first to investigate the comprehensive impact of directed topology in homophily and heterophily on the semi-supervised node classification paradigm, providing valuable empirical analysis for graph-based data engineering.
    (2) \textit{\underline{New Data Engineering Framework.}} 
    We introduce AMUD, which offers modeling guidance for natural digraphs from a statistical perspective, maximizing the benefits for subsequent graph learning. 
    (3) \textit{\underline{New Digraph Learning Paradigm.}} 
    We propose ADPA, which adaptively discovers the personalized DPs of each node by two hierarchical node-wise attention mechanisms, achieving effective message aggregation.
    This new paradigm introduces valuable insights for the future of digraph learning.
    (4) \textit{\underline{SOTA Performance.}}
    AMUD achieves a 4.57\% performance boost compared to ambiguous data engineering.
    Meanwhile, ADPA outperforms the most competitive baseline, achieving an average improvement of 4.16\% on directed modeling.

\section{Preliminaries}
\subsection{Notation and Problem Formalization}
\label{sec: preliminaries}
\noindent
    \textbf{Notation.}
    We consider a general graph representation method $G = (\mathcal{V}, \mathcal{E})$ with $|\mathcal{V}|=n$ nodes and $|\mathcal{E}|=m$ edges.
    The adjacency matrix (including self-loops) for the undirected graph is $\hat{\mathbf{A}}_u\in\mathbb{R}^{n\times n}$ and the directed variant of $\hat{\mathbf{A}}_u$ can be described by an asymmetrical matrix $\hat{\mathbf{A}}_d(u, v),u, v \in \mathcal{V}$. 
    $\hat{\mathbf{A}}_d(u, v)=1$ if $(u, v) \in \mathcal{E}$ and $\mathbf{A}_d(u, v)=0$ otherwise.
    For both undirected and directed scenarios, the node feature matrix is $\mathbf{X} = \{x_1,\dots,x_n\}$, where $x_v\in\mathbb{R}^{f}$ represents the feature vector of node $v$.
    Besides, $\mathbf{Y}$ is the label matrix.
    The semi-supervised node classification paradigm is based on the topology of labeled set $\mathcal{V}_L$ and unlabeled set $\mathcal{V}_U$, and the nodes in $\mathcal{V}_U$ are predicted based on the model supervised by $\mathcal{V}_L$.

\noindent
    \textbf{Problem Formalization with Our Proposal.}
    To achieve effective graph learning, we propose AMUD to guide the topological modeling of newly collected digraphs. 
    Subsequently, we can feed the undirected/directed output of AMUD (AMUndirected/AMDirected) into existing methods as shown in Fig.~\ref{fig: motivation_AMUD_AdpA}. 
    The motivation for this design is discussed in Sec.~\ref{sec: AMUD Guidance}.
    Although undirected GNNs have been extensively researched, the optimal learning paradigm for digraphs still lacks exploration. 
    Therefore, we propose ADPA specifically for AMDirected, which is also a feasible choice to be applied to AMUndirected thanks to its powerful capabilities.

\begin{figure}[t]   
	\centering
    \setlength{\abovecaptionskip}{0.2cm}
    \setlength{\belowcaptionskip}{-0.4cm}
	\includegraphics[width=\linewidth,scale=1.00]{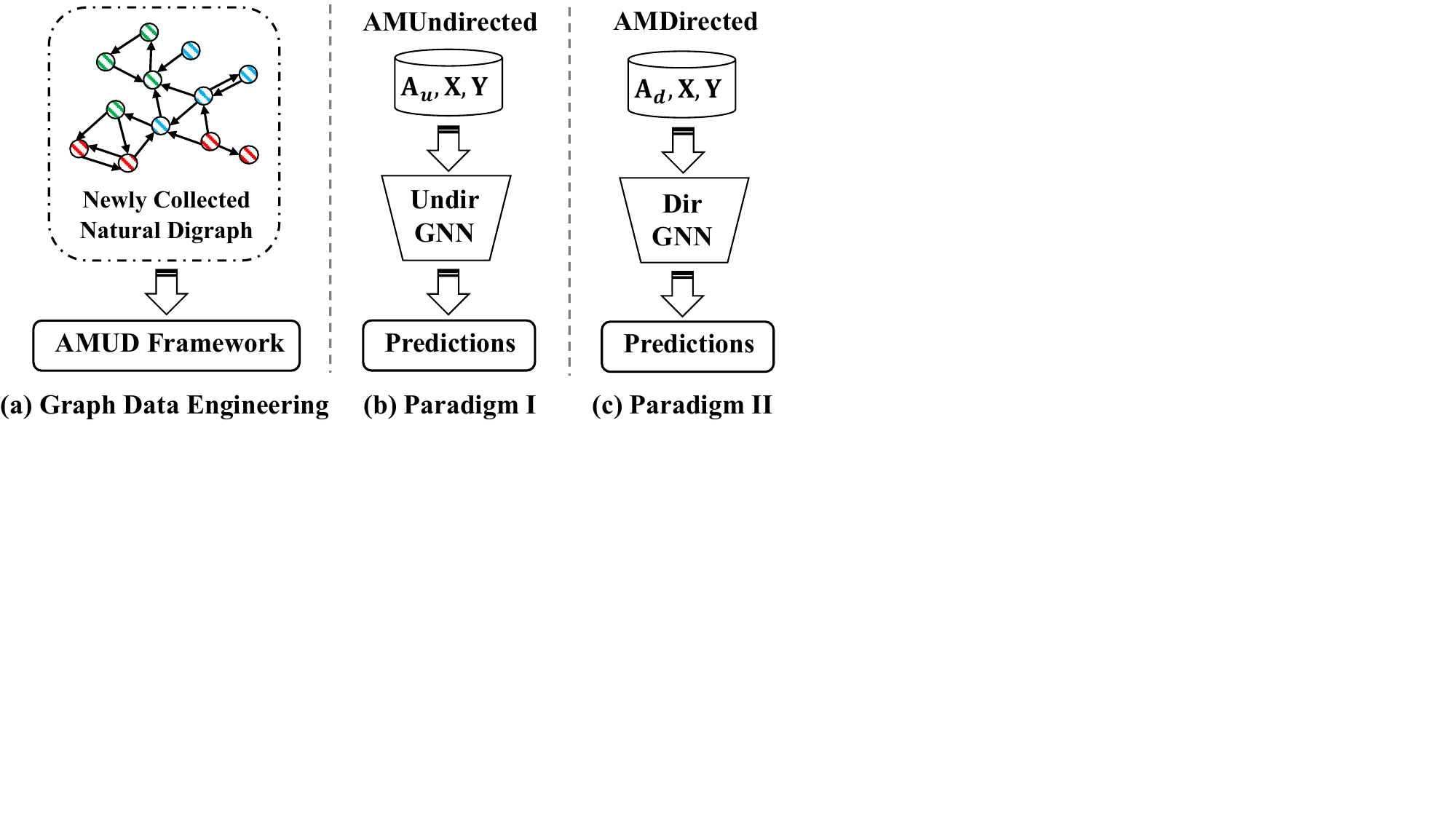}
     \captionsetup{font={small,stretch=1}}
	\caption{
    Workflow with our proposal.
    Paradigm I/II represents the dichotomy of the learning process determined by the output of AMUD.}
	\label{fig: motivation_AMUD_AdpA}
\end{figure}

\subsection{Undirected Graph Neural Networks}
\label{sec: Undirected Graph Neural Networks}
\noindent
    \textbf{Homophilous Methods.}
    Drawing inspiration from the spectral graph theory~\cite{chung2005spectral_graph_magnetic_laplacian1} and deep architectures, the most popular GCN~\cite{kipf2016gcn} is proposed, which can be viewed as a learnable first-order message aggregation. 
    From the perspective of design principles, this method aligns with the prominent homophily assumption~\cite{mcpherson200homophily_theory1,M2003Mixing_homophily_theory2,0Networks_homophily_theory3}, which finds an analog in the concept of smoothness or clustering assumption~\cite{10.5555/2968618.2968693_homophily_theory4} in the context of semi-supervised node classification. 
    Notably, the consistent and strong performance of label propagation observed across various datasets provides empirical support for enabling this iterative positive propagation of node features to neighbors during prediction.
    Formally, the $l$-th layer in GCN is
    \begin{equation}
        \label{eq: gcn}
        \mathbf{X}^{(l)}\! =\! \sigma(\tilde{\mathbf{A}}\mathbf{X}^{(l-1)}\mathbf{W}^{(l)}),\;\tilde{\mathbf{A}} \!=\! \hat{\mathbf{D}}^{r-1}\hat{\mathbf{A}}\hat{\mathbf{D}}^{-r},\;r\in[0,1],
    \end{equation}
    where $\hat{\mathbf{D}}$ is the degree matrix of $\hat{\mathbf{A}}$, $r$ is the convolution coefficient, $\mathbf{W}$ is the trainable weights, and $\sigma(\cdot)$ is the non-linear activation function.
    By selecting appropriate $r$, we obtain the random walk-based $\hat{\mathbf{A}}\hat{\mathbf{D}}^{-1}$\cite{xu2018jknet}, symmetric normalized $\hat{\mathbf{D}}^{-1/2}\hat{\mathbf{A}}\hat{\mathbf{D}}^{-1/2}$, and the reverse transition-based $\hat{\mathbf{D}}^{-1}\hat{\mathbf{A}}$\cite{zeng2019graphsaint}.
    By utilizing them, recent studies~\cite{hamilton2017graphsage,2019appnp,wang2020gcnlpa,chen2020gcnii, Hu2021ahgae} optimize the model architectures to improve performance. 

\noindent
    {
    \textbf{Heterophilous Methods.}
    Despite homophilous GNNs' effectiveness, recent surveys~\cite{ma2021hete_gnn_survey1,luan2022hete_gnn_survey2,zheng2022hete_gnn_survey3} reveal the limitations of directly deploying these methods. 
    Specifically, naive homophilous propagation can be viewed as feature augmentation.
    However, heterophily misleads this process and disturbs the node representations.
    Recent approaches aim to capture heterophily by incorporating high-order neighbors or multi-scale messages~\cite{luo2022ambiguous_graph_de_icde5,lv2023datairreversible_graph_vldb5,wang2023scapin_sigmod5}.
    H$_2$GCN introduces neighbor rules, such as $\tilde{\mathbf{A}}_1$ and $\tilde{\mathbf{A}}_2$, to obtain $\mathbf{Z}\!=\!\operatorname{Combine}{(\operatorname{Agg}(\tilde{\mathbf{A}}_1,\mathbf{X}), \operatorname{Agg}(\tilde{\mathbf{A}}_2,\mathbf{X}))}$.
    GPR-GNN~\cite{chien2021gprgnn} controls the contribution of propagated features in each step by learnable PageRank weights $\mathbf{Z}=\sum_{k=0}^K \gamma_k \mathbf{X}^{(k)}$.
    LINKX~\cite{2021linkx} separately encodes node features and topology, avoiding negative impacts caused by misleading interactions $\mathbf{Z}\! = \!\operatorname{Concat}(\operatorname{MLP}(\mathbf{X}),\operatorname{MLP}(\mathbf{A}))$.
    GloGNN~\cite{2022glognn} further utilizes global transformation probability matrix $\mathbf{T}$ to generates a node’s global embedding $\mathbf{Z}=(1-\gamma)\mathbf{T}^{(l)}\mathbf{X}^{(l)}+\gamma \mathbf{X}^{(l)}$.
    Inspired by them, several recent approaches~\cite{pei2020geomgcn,yan2021ggcn,dai2022lwgcn,du2022gbkgnn,song2023ordergnn} further improve predictive performance through well-designed model architectures and optimized aggregation strategies.}

\noindent
    {\textbf{Homophily Measures.}
    Despite the considerable efforts of homophilous or heterophilous GNNs, quantifying topological homophily remains a challenge~\cite{zeng2022ambiguous_graph_de_icde4,lv2023henceirreversible_graph_vldb4,wan2023scalable_sigmod4}. 
The widely used metrics are node homophily $\operatorname{H}_{node}$~\cite{pei2020geomgcn} and edge homophily $\operatorname{H}_{edge}$~\cite{2020h2gcn}, which compute the proportion of edge-connected nodes and nodes' neighbors that share the same class, respectively.
    These measures are straightforward and intuitive. 
    However, they exhibit sensitivity to the label distributions, which hinders their reliability.
    In response to these limitations, $\operatorname{H}_{class}$~\cite{2021linkx}, $\operatorname{H}_{adj}$~\cite{platonov2023hete_gnn_survey4}, and $\operatorname{LI}$~\cite{platonov2022hete_gnn_survey5} are proposed.
    Despite theoretical analysis indicates that these metrics exhibit stronger robustness, they are still constrained by the representation challenge posed by the sub-optimal undirected topology.}

\subsection{Directed Graph Neural Networks}

\noindent
    \textbf{Spatial-based Methods.}
    To obtain node embeddings in digraphs with asymmetrical topology $\mathbf{A}_d$, spatial-based methods follow the message-passing mechanism in undirected scenarios. 
    However, it's crucial to consider the directed edges when aggregating messages $\operatorname{Agg}(\cdot)$. 
    Thus, current node $i \in \mathcal{V}$ adopts independent learnable weights over out-neighbors $(i\rightarrow j)$ and in-neighbors $(j\rightarrow i)$ to combine representation $\operatorname{Com}(\cdot)$.
    \begin{equation}
    \begin{aligned}
        \label{eq: spatial_dignn}
        \mathbf{H}_{i,\rightarrow}^{(l)}&=\operatorname{Agg}\left(\mathbf{X}_i^{(l-1)},\mathbf{X}_j^{(l-1)},\{\forall (i,j)\in\mathcal{E}\}\right),\\
        \mathbf{H}_{i,\leftarrow}^{(l)}&=\operatorname{Agg}\left(\mathbf{X}_i^{(l-1)},\mathbf{X}_j^{(l-1)},\{\forall (j,i)\in\mathcal{E}\}\right),\\
        &\mathbf{X}_i^{(l)} = \operatorname{Com}\left(\mathbf{X}_i^{(l-1)}, \mathbf{H}_{i,\leftarrow}^{(l)}, \mathbf{H}_{i,\rightarrow}^{(l)}\right).
        \end{aligned}
    \end{equation}
    Building upon this foundation, DGCN~\cite{tong2020dgcn} introduces first and second-order neighbor proximity to devise message aggregation strategies.
    DIMPA~\cite{he2022dimpa} increases the receptive field of nodes by aggregating $K$-hop neighborhoods at each layer.
    NSTE~\cite{kollias2022nste} is inspired by the 1-WL graph isomorphism test, tuning based on the directed propagation.
    DiGCN~\cite{tong2020digcn} leverages the neighbor proximity to increase the receptive field and theoretically extends personalized PageRank to construct a real symmetric digraph Laplacian.
    This method advances in extending undirected spectral convolution to digraph scenarios.
    
\noindent
    \textbf{Spectral-based Methods.}
    To implement spectral convolution on digraphs with theoretical guarantees, the core is to obtain a symmetric (conjugated) digraph Laplacian $\mathbf{L}_d$ based on $\mathbf{A}_d$.
    \begin{equation}
    \begin{aligned}
        \label{eq: spectral_dignn}
        &\;\;\mathbf{L}_d = \operatorname{DGS}(\mathbf{A}_d, \alpha, q),\\
        \hat{\mathbf{Y}} = &\operatorname{MLP}\left(\operatorname{Poly}\left(\mathbf{L}_d\right)\operatorname{MLP}\left(\mathbf{X}\right)\right),
        \end{aligned}
    \end{equation}
    where $\operatorname{DGS}(\cdot)$ is the digraph generalized symmetric function with parameters and $\hat{\mathbf{Y}}$ being the predictions.
    Building upon this, DiGCN~\cite{tong2020digcn} introduces the $\alpha$-parameterized stable state distribution based on the personalized PageRank to achieve digraph convolution.
    MagNet~\cite{zhang2021magnet} utilizes $q$-parameterized complex Hermitian matrix (magnetic Laplacian) to define convolution with independent trainable weights for the real and image parts.
    MGC~\cite{zhang2021mgc} adopts a truncated variant of PageRank on magnetic Laplacian for fine-grained filtering.
    Furthermore, $\operatorname{Poly}(\cdot)$ is a polynomial-based approximation method, such as the Linear-Rank~\cite{baeza2006linear_rank} adopted in MGC and the first-order Chebyshev polynomial employed in MagNet and DiGCN.

\section{AMUD: Graph-based Data Engineering}
\label{sec: AMUD: Graph-based Data Engineering}
\subsection{Empirical Analysis}
\label{sec: Empirical Analysis}
    To illustrate the two empirical observations introduced in Sec.~\ref{sec: introduction}, we provide an intuitive description on the left side of Fig.~\ref{fig: motivation_observation} and corresponding experimental results on the right side. 
    This demonstrates our concerns regarding the ambiguous integration of graph-based data engineering and graph learning process and provides ample support for our following claims.

    {To answer (\textbf{O1}), we provide experimental results as shown in Fig.~\ref{fig: motivation_observation}(a) and (b). 
    We can observe that feeding the naturally directed CoraML into undirected GNNs after coarse undirected transformation results in better predictive performance compared to directly feeding it into directed GNNs. 
    However, this phenomenon is entirely the opposite for the Chameleon dataset. 
    To investigate the reasons behind this occurrence from the perspective of topology, we quantify the homophily using the metrics introduced in Sec.~\ref{sec: Undirected Graph Neural Networks}, as shown in Table~\ref{tab: emprical_topology}.
    To explain it, we propose Proposition~\ref{proposition: for_observation1}, which aligns with our results and further investigates this issue by Question~\ref{question: for_observation1}. 
    Notably, the conclusion regarding undirected graphs with homophily has been confirmed in recent related works~\cite{song2022gnn_survey4}, but the relationship between digraphs and topological properties lacks investigation.
    \begin{proposition}
    \label{proposition: for_observation1}
    \!Undirected GNNs are more suitable for handling homophilous undirected graphs, while directed GNNs exhibit a significant advantage in dealing with heterophilous digraphs.
    \end{proposition}
    \begin{question}
    \label{question: for_observation1}
    \!What does modeling directed information mean for directed GNNs, and whether it aid in capturing homophily or heterophily during the graph learning process?
    \end{question} 
    }

    {To further answer Question~\ref{question: for_observation1} and explore the reasons behind (\textbf{O2}), we present experimental results in Fig.~\ref{fig: motivation_observation}(c) and (d). 
    It is evident that, for directed GNNs, undirected augmentation (i.e., directly converting directed edges into undirected edges) on CiteSeer leads to a significant performance boost, while it has a negative impact on Squirrel.
    Considering the statistical information in Table~\ref{tab: emprical_topology}, we propose Proposition~\ref{proposition: for_observation2}.
    This indicates that directed information provides us with a new perspective to dissect the entanglement of homophily and heterophily in the conventional undirected graph learning process.
    \begin{proposition}
    \label{proposition: for_observation2}
    \!Modeling directed information assists directed GNNs in capturing intricate heterophily while discarding directed information is more crucial for utilizing homophily.
    \end{proposition}}

    {It is well-known that in recent years, numerous metrics for quantifying homophily have been proposed.
    A natural idea is to directly employ these metrics to determine whether to model directed topology. 
    However, existing homophilous measures ignore directed information, making them sub-optimal choices in digraph scenarios.
    To address this issue, we propose a unified data engineering framework, AMUD, from a statistical perspective.
    It represents the correlation between homophily and directed edges through linear regression and calculates the coefficient of determination ($R^2$) to guide the modeling rules for newly collected digraphs, maximizing the benefits in subsequent undirected or directed graph learning processes.}

\begin{figure*}[t]
  \includegraphics[width=\textwidth]{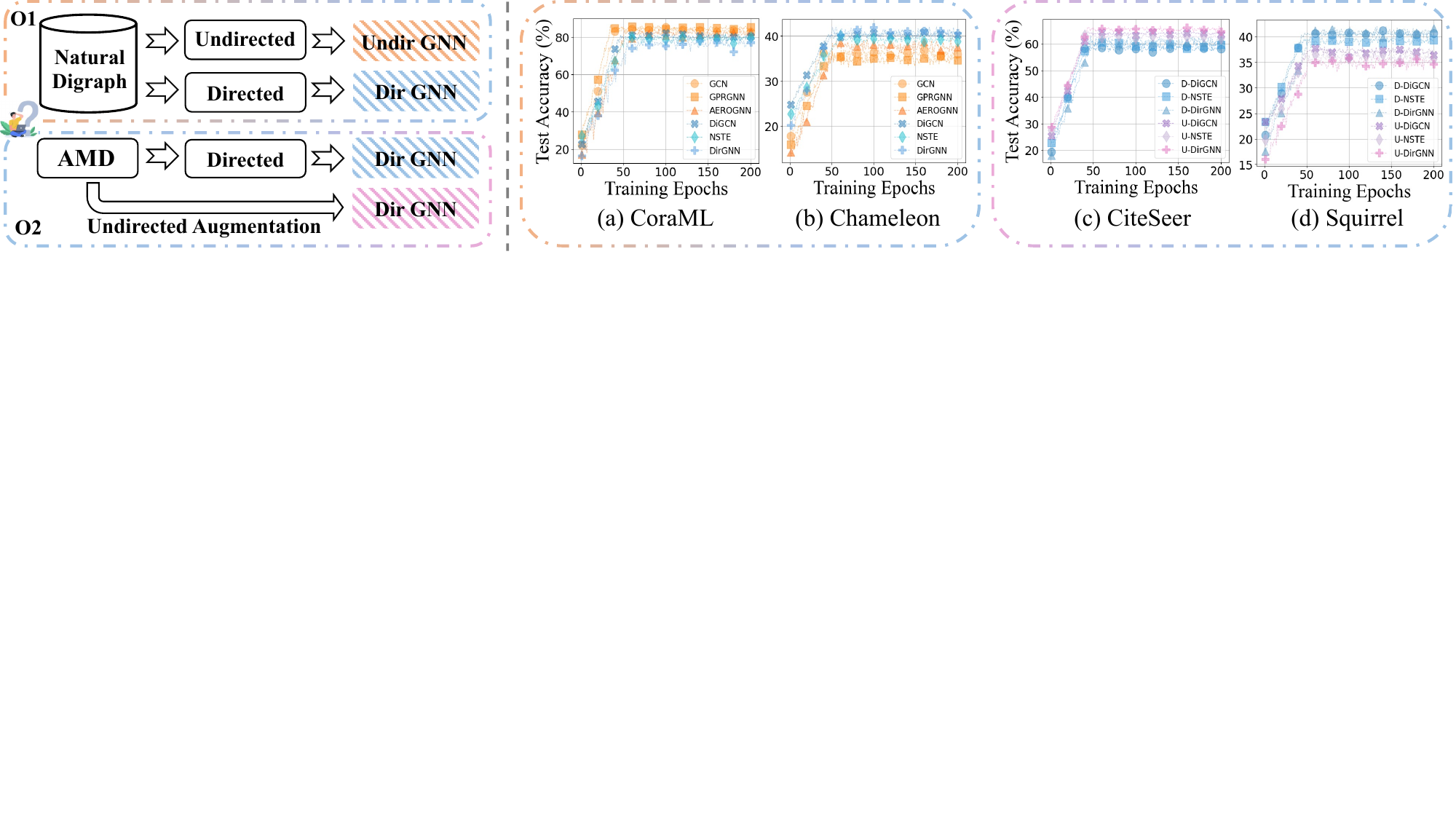}
      \captionsetup{font={small,stretch=1}}
\setlength{\abovecaptionskip}{-0.2cm}
\setlength{\belowcaptionskip}{-0.2cm}
  \caption{
    The two observations mentioned in Sec.~\ref{sec: introduction} \textbf{L2}.
    AMD is the directed output of AMUD.
    CoraML, Chameleon, CiteSeer, and Squirrel are four natural digraphs, where Chameleon and Squirrel are filter versions in~\cite{platonov2023hete_gnn_survey4}.
    GCN, GRP-GNN, and AEROGNN are three undirected GNNs. 
    DiGCN, NSTE, and DirGNN are three directed GNNs.
    U- and D- represent the input of undirected and directed graphs.}
  \label{fig: motivation_observation}
\end{figure*}

\begin{table}[t]
\setlength{\abovecaptionskip}{0.2cm}
\setlength{\belowcaptionskip}{0cm}
\caption{Homophily from naturally \sethlcolor{red!20}\hl{directed} to coarse \sethlcolor{blue!15}\hl{undirected} transformation and our proposed AMUD for the \textbf{directed scenario}.
}

\footnotesize
\label{tab: emprical_topology}
\resizebox{\linewidth}{9.5mm}{
\setlength{\tabcolsep}{0.8mm}{
\begin{tabular}{c|cccccc}
\hline
Datasets  & $\operatorname{H}_{node}$ & $\operatorname{H}_{edge}$ & $\operatorname{H}_{class}$ & $\operatorname{H}_{adj}$ & $\operatorname{LI}$ & AMUD \\ \hline
CoraML    & \sethlcolor{red!20}\hl{.792}-\sethlcolor{blue!15}\hl{.789}                       & \sethlcolor{red!20}\hl{.808}-\sethlcolor{blue!15}\hl{.810}                       & \sethlcolor{red!20}\hl{.744}-\sethlcolor{blue!15}\hl{.740}                        & \sethlcolor{red!20}\hl{.784}-\sethlcolor{blue!15}\hl{.780}                      & \sethlcolor{red!20}\hl{.567}-\sethlcolor{blue!15}\hl{.561}                 & 0.380    \\
Chameleon & \sethlcolor{red!20}\hl{.245}-\sethlcolor{blue!15}\hl{.236}                       & \sethlcolor{red!20}\hl{.247}-\sethlcolor{blue!15}\hl{.244}                       & \sethlcolor{red!20}\hl{.058}-\sethlcolor{blue!15}\hl{.044}                        & \sethlcolor{red!20}\hl{.203}-\sethlcolor{blue!15}\hl{.193}                      & \sethlcolor{red!20}\hl{.027}-\sethlcolor{blue!15}\hl{.014}                 & 0.657    \\
CiteSeer  & \sethlcolor{red!20}\hl{.739}-\sethlcolor{blue!15}\hl{.738}                       & \sethlcolor{red!20}\hl{.725}-\sethlcolor{blue!15}\hl{.720}                       & \sethlcolor{red!20}\hl{.627}-\sethlcolor{blue!15}\hl{.629}                        & \sethlcolor{red!20}\hl{.726}-\sethlcolor{blue!15}\hl{.724}                      & \sethlcolor{red!20}\hl{.475}-\sethlcolor{blue!15}\hl{.454}                 & 0.269    \\
Squirrel  & \sethlcolor{red!20}\hl{.216}-\sethlcolor{blue!15}\hl{.207}                       & \sethlcolor{red!20}\hl{.185}-\sethlcolor{blue!15}\hl{.191}                       & \sethlcolor{red!20}\hl{.067}-\sethlcolor{blue!15}\hl{.040}                        & \sethlcolor{red!20}\hl{.173}-\sethlcolor{blue!15}\hl{.165}                    & \sethlcolor{red!20}\hl{-.014}-\sethlcolor{blue!15}\hl{.001}                & 0.693    \\ \hline
\end{tabular}
}}
\vspace{-0.2cm}
\end{table}

\begin{figure}[t]   
	\centering
    \setlength{\abovecaptionskip}{0.2cm}
    \setlength{\belowcaptionskip}{-0.5cm}
	\includegraphics[width=\linewidth,scale=1.00]{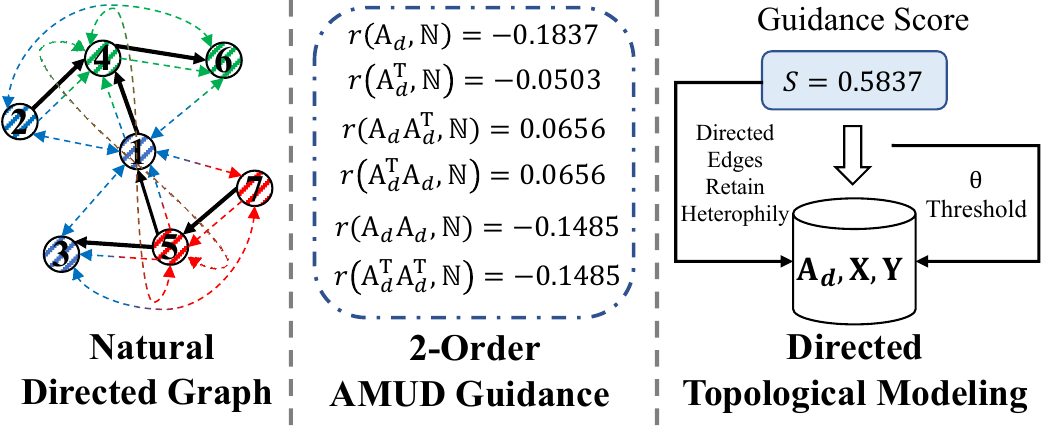}
     \captionsetup{font={small,stretch=1}}
	\caption{A toy example for our proposed AMUD.}
	\label{fig: amud_toy_example}
\end{figure}

\subsection{AMUD Formalization}
    As discussed in Sec.~\ref{sec: introduction} and Sec.~\ref{sec: Empirical Analysis}, the entanglement of homophily and heterophily is concealed beneath the directed edges. 
    To validate our claims, we present relevant statistics in Table~\ref{tab: emprical_topology}, including the homophily (CoraML and CiteSeer) and heterophily (Chameleon and Squirrel) which has been generally recognized by previous studies~\cite{platonov2023hete_gnn_survey4,platonov2022hete_gnn_survey5}.
    Building upon this, we aim to highlight the advantages of AMUD over other metrics.
    According to Table~\ref{tab: emprical_topology}, we observe that existing metrics exhibit inefficiency when confronted with directed topologies. 
    Specifically, they fail to distinctly reveal differences in the same dataset under directed and undirected scenarios. 
    For instance, the disparities in the undirected and directed versions of homophilous CoraML and CiteSeer under $\operatorname{H}_{edge}$ and $\operatorname{H}_{adj}$ are nearly negligible. Meanwhile, for the heterophilous Chameleon and Squirrel, the differences only remain modest, e.g., \sethlcolor{red!20}\hl{.245} to \sethlcolor{blue!15}\hl{.236} and \sethlcolor{red!20}\hl{.173} to \sethlcolor{blue!15}\hl{.165}, respectively.
    
    To address this issue, we propose AMUD tailored for directed scenarios, which establishes a statistical connection between nodes and directed edges to guide graph-based data engineering.
    A toy example is shown in Fig.~\ref{fig: amud_toy_example}.
    To begin with, we emphasize the importance of directed topology in preserving intricate relationships, highlighting the information loss caused by coarse undirected transformation.
    For Node 1, \sethlcolor{blue!35}\hl{Node 2} and \sethlcolor{blue!35}\hl{Node 3} share the same node class assisted by the same relationship in the out-neighbor ($\rightarrow$\sethlcolor{green!40}\hl{Node 4}) and in-neighbor ($\leftarrow$\sethlcolor{red!45}\hl{Node 5}) of Node 1. 
    However, \sethlcolor{green!45}\hl{Node 6} and \sethlcolor{red!45}\hl{Node 7} have different classes caused by the different relationship in the neighbors of Node 1 (i.e., $\rightarrow,\leftarrow$\sethlcolor{green!40}\hl{Node 4} and $\leftarrow,\rightarrow$\sethlcolor{red!45}\hl{Node 5}).
    This critical topological information is lost when simply transforming digraphs into undirected ones, which results in Node 1 integrating all of its 2-hop neighbors, entangling the homophily and heterophily.

    For fine-grained quantification of topological characterizes (i.e., homophily and heterophily) concealed beneath directed edges, we present the derivation and computation details of AMUD from a statistical perspective.
    It is worth noting that, for simplicity, we utilize $\mathbf{G}_d$ to represent different high-order neighbors, such as 1-hop neighbors, 2-hop neighbors, and so on. 
    That is, for any $u,v \in \mathcal{V}$, $\mathbf{G}_d(u,v)=1$ if $u,v$ are high-order neighbors (maybe not directly connected) and $\mathbf{G}_d(u,v)=0$ otherwise. 
    Building upon this, we have the topology $\mathbf{G}_d$ and node profiles $\mathbb{N}$ (e.g., features or labels) of a given digraph.
    Let a random variable $\mathbf{G}_d$ be the multi-scale relationships of the randomly picked directed topology in $\mathbf{A}_d$. 
    The distribution of $\mathbf{G}_d$ is $\mathbf{P}({\mathbf{G}_d}(u,v)) = \mathbf{P}({\mathbf{G}_d}(u,v) = G_d(\mathcal{V}_d,\mathcal{E}_d))$, where $G_d(\mathcal{V}_d,\mathcal{E}_d)$ is the outcome digraph when we pick multi-scale relationships $\mathbf{G}_d$.
    Similarly, let a random variable $\mathbb{N}$ be a node profile associated with the randomly picked comprehensive description in $\mathcal{V}(\mathbf{X}, \mathbf{Y})$, the distribution of $\mathbb{N}$ is $\mathbf{P}\left(\mathbb{N}(u)\right)=\mathbf{P}(\mathbb{N}(u)=\mathcal{V}_u(x_u,y_u))$, where $\mathcal{V}_u(x_u,y_u)$ is the outcome description when we pick node u’s node profile $\mathbb{N}(u)$. 
    
    \begin{equation}
        \begin{aligned}
            \label{eq: probability_distribution}
            \mathbf{P}\left(\mathbf{G}_d(u,v)\right) &= \frac{\mathbf{1}_{\mathbf{G}_d(u,v) = k, k\in \{ 0,1 \} }}{\binom{n}{2}},\\
            \mathbf{P}\left(\mathbb{N}(u)\right) &= \frac{\mathbf{1}_{\mathbb{N}(u) = \mathbf{N}, \mathbf{N}\in \mathbb{R}^{\mathbb{N}}}}{n},
        \end{aligned}
    \end{equation}

    where $\mathbf{1}_{\mathbf{G}_d(u,v) = k, k\in \{ 0,1 \} }$ and $\mathbf{1}_{\mathbb{N}(u) = \mathbf{N}, \mathbf{N}\in \mathbb{R}^{\mathbb{N}}}$ represent the numbers of elements that equal to  $k\in \{ 0,1 \}$ and $\mathbf{N}\in \mathbb{R}^{\mathbb{N}}$, respectively. 
    Then, we calculate the expectations of them.
    \begin{equation}
        \begin{aligned}
            \label{eq: expectation}
            \mathbb{E}(\mathbf{G}_d) &= \sum_{u,v \in \mathcal{V}} \mathbf{G}_d(u,v)\times \mathbf{P}(\mathbf{G}_d(u,v)),\\
            \mathbb{E}(\mathbb{N}) &= \sum_{u\in \mathcal{V}} \mathbb{N}(u)\times \mathbf{P}(\mathbb{N}(u)).
        \end{aligned}
    \end{equation}
    
    After that, we calculate the covariance between directed topology $\mathbf{G}_d$ and node description $\mathbb{N}$ to quantify the relevance between multi-scale relationships and node profiles.
    \begin{equation}
        \begin{aligned}
            \label{eq: covariance}
            \operatorname{Cov}(\mathbf{G}_d, \mathbb{N}) &= \mathbb{E}((\mathbf{G}_d - \mathbb{E}(\mathbf{G}_d))\times (\mathbb{N} - \mathbb{E}(\mathbb{N}))) \\
                                        &= \mathbb{E}(\mathbf{G}_d\mathbb{N})-\mathbb{E}(\mathbf{G}_d)\mathbb{E}(\mathbb{N}).
        \end{aligned}
    \end{equation}

    Based on the above equations, we calculate the Pearson correlation coefficient $r(\mathbf{G}_d,\mathbb{N})$ of digraph operator $\mathbf{G}_d$ and attribute matrix $\mathbb{N}$, which formally defined as follows:
    \begin{equation}
        \begin{aligned}
            \label{eq: coefficient}
            \frac{\operatorname{Cov}(\mathbf{G}_d, \mathbb{N})}
            {\sqrt{\sum_{u,v\in \mathcal{V}}(\mathbf{G}_d(u,v)-\overline{\mathbf{G}_d})^2}
            \sqrt{\sum_{u\in \mathcal{V}}(\mathbb{N}(u)-\overline{\mathbb{N}})^2}},
        \end{aligned}
    \end{equation}

    \noindent
    where $\overline{\mathbf{G}_d}$ and $\overline{\mathbb{N}}$ are the mean value $\mathbf{G}_d$ and $\mathbb{N}$. 
    Notably, for a linear model of $\mathbf{G}_d$ and $\mathbb{N}$, the square of $r(\mathbf{G}_d,\mathbb{N})$ is denoted as $R^2{(\mathbf{G}_d,\mathbb{N})}$. 
    This metric signifies that a coefficient closer to 1 implies a stronger linear relation between directed relationships and node profiles.
    The criteria for deciding whether to perform directed modeling are introduced subsequently.

\subsection{Directed Modeling Guidance}
\label{sec: directed modeling guidance}
    Based on the above definition, $R^2{(\mathbf{G}_d,\mathbb{N})}$ can be extended by considering higher-order relationships $\mathbf{G}_d$ and more comprehensive node descriptions $\mathbb{N}$. 
    However, in our implementation, we quantify the correlations between 2-hop neighbors and node labels for enhancing efficiency.
    Notably, this strategy is enough to capture abundant semantic information akin to compact motifs in complex networks, recognized as powerful tools in graph mining~\cite{milo2002_network_motif_1, benson2016_network_motif_2, ribeiro2021_network_motif_3}.
    For instance, ${\mathbf{A}}_d{\mathbf{A}}_d^T$ and ${\mathbf{A}}_d^T{\mathbf{A}}_d$ provide more abundant homophily than other 2-order DPs, which implies that these 2-hop neighbors are more likely to share similar features and the same labels with the current node.
    In contrast, ${\mathbf{A}}_d{\mathbf{A}}_d$ and ${\mathbf{A}}_d^T{\mathbf{A}}_d^T$ provide support for modeling intricate heterophilous relationships.
    This key insight prompts us to reveal topological relationships concealed within directed edges by measuring the correlation between $\mathbf{G}_d$ and $\mathbb{N}$.

    Specifically, if the disparity in correlations between DPs and node profiles exceeds the threshold, it indicates the necessity of retaining directed topology to reveal intricate relationships. 
    Otherwise, undirected transformation is recommended.
    Formally, we define the guidance score $S$ by 

\begin{equation}
    \begin{aligned}
        \label{eq: guidance_score}
        S \!&= \!\alpha\sqrt{\sum_{{\mathbf{G}_d}_i,{\mathbf{G}_d}_j,{\mathbf{G}_d}_i\neq{\mathbf{G}_d}_j} \!\!\!\!\left\Vert  R^2{({\mathbf{G}_d}_i,\mathbb{N})}-R^2{({\mathbf{G}_d}_j,\mathbb{N})} \right\Vert_2} / \binom{4}{2}.\\
    \end{aligned}
\end{equation}

\noindent
    Given the sparsity of real-world digraphs, $R^2{(\mathbf{G}_d,\mathbb{N})}$ tends to be a small value. 
    To address this, we introduce a scalar operator $\alpha=1/\max\left(R^2{(\mathbf{G}_d,\mathbb{N})}\right)$. 
    In our implementation, if $S>\theta=0.5$, we should retain its directed edges. 
    Otherwise, it is recommended to model it into an undirected graph.

\begin{figure*}[t]
  \includegraphics[width=\textwidth]{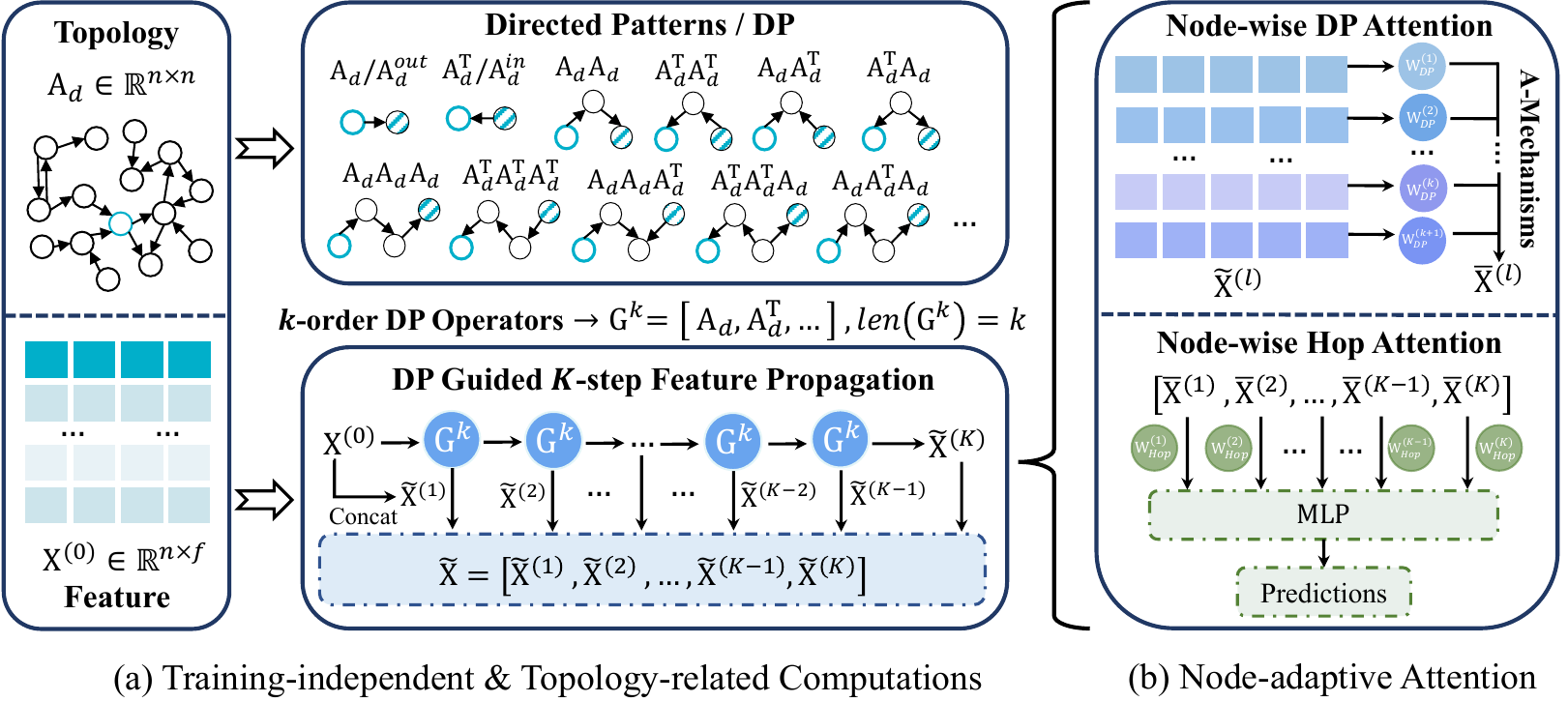}
  \captionsetup{font={small,stretch=1}}
  \caption{
  Overview of our proposed ADPA, including 
  (a) discover DPs and achieve multi-scale directed feature propagation; 
  (b) combine the $K$-step propagated features within $k$-order DPs to obtain multi-granularity node representations by node-adaptive attention.}
  \label{fig: model_framework}
  \vspace{-0.2cm}
\end{figure*}

\section{Adaptive Directed Pattern Aggregation}
\label{sec: Adaptive Directed Pattern Aggregation}
    Building upon the directed topological modeling guidance, as illustrated in Fig.~\ref{fig: motivation_AMUD_AdpA}, we suggest that feeding AMUndirected into existing state-of-the-art undirected GNNs. 
    This is because, in contrast to directed GNNs, these methods focus exclusively on undirected inputs and exhibit performance advantages due to well-designed propagation strategies.
    Moreover, many powerful undirected GNNs have been proposed in recent years, showcasing sufficient capacity to handle such undirected scenarios. 
    In contrast, a consensus on the optimal digraph learning paradigm for AMDirected is yet to be established.
    
    To address this issue, we propose ADPA shown in Fig.~\ref{fig: model_framework}.
    In this section, we commence by introducing the key intuitions of our proposal in Sec.~\ref{sec: Architecture Overview}.
    Subsequently, in Sec.~\ref{sec: directed patterns guided feature propagation} and Sec.~\ref{sec: Node-adaptive Attention Mechanisms}, we elaborate on the details of ADPA, which utilizes adaptive exploration of diverse DPs to extract multi-scale directed information and incorporates two hierarchical node-adaptive attention mechanisms to facilitate the integration of multi-granularity fusion representations. 
    Meanwhile, we provide the algorithm complexity analysis of ADPA in Sec.~\ref{sec: complexity analysis}. 
    Notably, serving as a new digraph learning paradigm, ADPA provides a unified framework for existing methods that adhere to the message-passing framework. 
    Based on this, we illustrate that most of the existing spatial-based approaches can be considered a specific instance of ADPA in Sec.~\ref{sec: Comparison between the Spatial-based GNNs and ADPA}.
    To provide a comprehensive understanding of our proposed workflow, we present detailed illustrations in Alg.~\ref{alg: amud_adpa}.

\subsection{Architecture Overview}
\label{sec: Architecture Overview}
    According to the analysis in Sec.~\ref{sec: introduction} and Sec.~\ref{sec: AMUD: Graph-based Data Engineering}, we observe that the directed topology introduces intricate relationships among nodes in the digraph. 
    This complexity is beyond the commonly acknowledged one-hop neighbor relationships in undirected scenarios, evolving into the exploration of complex topological co-occurrences in higher-order neighbors. 
    Moreover, label-guided homophily and heterophily also play a significant role in this scenario.
    Therefore, we aim to propose a unified framework that can effectively handle directed topology while breaking the entanglement of homophily and heterophily in the semi-supervised node classification paradigm.
    The following content elucidates the intuition of our proposed ADPA.

    \underline{\emph{Directed Pattern Guided Feature Propagation.}} 
    As depicted in Table~\ref{tab: emprical_topology} and Fig.~\ref{fig: amud_toy_example}, the intricate homophily and heterophily in directed scenarios significantly differ from the undirected ones. 
    In digraphs, the co-occurrence patterns in the topological structure, enriched by the fusion of directed information, contain substantial yet undiscovered knowledge.
    To capture this intricate multi-scale structural insights of each node from both local and global perspectives, we extract various DPs into instantiated matrices to guide directed $K$-step feature propagation.
    Notably, to ensure computational efficiency, we discarded the tightly coupled frameworks in graph learning. 
    Instead, we introduce a weight-free feature propagation module independent of the training. 
    Similar design principles have been proven to significantly enhance computational efficiency without compromising predictive performance~\cite{wu2019sgc,frasca2020sign,sun2021sagn,zhu2021ssgc,gamlp}.

    \underline{\emph{Node-adaptive Attention Mechanisms.}}
    After obtaining node representations that consider multi-scale structural information, a natural question arises: \textit{Whether the diverse DP-guided node representations and the propagated features at each step will collectively contribute positively to the final prediction?}
    Based on the conclusions drawn in Sec.~\ref{sec: AMUD: Graph-based Data Engineering} and relevant studies about graph propagation~\cite{2019appnp,zhang2021rod,chien2021gprgnn}, different DPs and propagation steps have diverse impacts on various datasets, leading to distinct influences on individual nodes. 
    Therefore, we propose two hierarchical node-wise attention mechanisms to finely fusion multi-granularity node representations.

    \underline{\emph{A New Digraph Learning Paradigm.}} 
    Notably, our proposed ADPA essentially serves as an instantiation of a unified framework for digraph learning. 
    We strive for ADPA to offer essential insights for subsequent studies based on message-passing, particularly focusing on the spatial domain propagation perspective.
    Specifically, ADPA can benefit from advancements in well-designed feature propagation strategies (e.g., initial residuals and dense connection) to obtain a more powerful multi-scale structural representation. 
    Meanwhile, researchers have the flexibility to replace attention components within the node-adaptive multi-granularity representation fusion with other reasonable approaches (e.g., gate attention and self-attention), enabling customization based on specific requirements.


\subsection{Directed Patterns Guided Feature Propagation}
\label{sec: directed patterns guided feature propagation}
    As illustrated in Sec.~\ref{sec: directed modeling guidance}, different DP operators provide different structural insights for various semantic information (e.g., intricate homophily and heterophily in digraphs). 
    This key insight motivates us to propose the various DPs to capture multi-scale structural information, facilitating information interaction between higher-order nodes by directed feature propagation.
    Specifically, given the directed topology $\mathbf{A}_d$ and node feature matrix $\mathbf{X}$, ADPA first generates $k$-order DPs to propagate the current node features to their multi-directed high-order neighbors.
    Formally, we extend Eq.~(\ref{eq: spatial_dignn}) by letting $\textbf{G}_d = \left \{ \mathbf{A}_d, \mathbf{A}_d^T, \mathbf{A}_d\mathbf{A}_d, \mathbf{A}_d^T\mathbf{A}_d^T,\cdots \right \}$ be the DP operators of length $k$ ($\mathbf{A}_d=\mathbf{A}_u$ in the undirected scenario), then the $K$-step parameter-free feature propagation is defined as
\begin{equation}
    \label{eq: k-order DP operator}
    \begin{aligned}
        &\tilde{\mathbf{X}}\!\!=\!\!\left[\tilde{\mathbf{X}}^{(1)},\dots,\tilde{\mathbf{X}}^{(K)}\right],\tilde{\mathbf{X}}^{(l)} \!\!=\!\! \left[\mathbf{X}^{(0)},\mathbf{X}_{\mathbf{G}_{d_1}}^{(l)},\dots ,\mathbf{X}_{\mathbf{G}_{d_k}}^{(l)}\right],\\
        &\mathbf{X}_{\mathbf{G}_{d_g}}^{(l)}\!\!=\mathbf{G}_{d_g}\mathbf{X}_{\mathbf{G}_{d_g}}^{(l-1)}, \forall g=1,\dots,k,\;\forall l=1,\dots,K,\\
        &\mathbf{X}_{i,\mathbf{G}_{d_g}}^{(l)}\!\!=\operatorname{Agg}\left(\mathbf{X}_{i,{\mathbf{G}_{d_g}}}^{(l-1)},\mathbf{X}_{j,{\mathbf{G}_{d_g}}}^{(l-1)},\{\forall i,j, \mathbf{G}_{d_g}(i,j)\!=\!1\}\right),
    \end{aligned}
\end{equation}
    where $\mathbf{X}^{(0)}=\mathbf{X}$, this initial residual has been demonstrated to be theoretically effective in recent studies~\cite{chen2020gcnii,zhang2022air}, from both spatial and spectral perspectives.
    Intuitively, this strategy can emphasize the unique attributes of each node during the feature propagation process, thereby alleviating the notorious over-smoothing issues and improving predictive performance.

    In our general implementation of Eq.~(\ref{eq: k-order DP operator}), the selection of $k$ follows specific rules. 
    When considering only the 1-hop neighborhood, $k$ is set to 2. 
    This is because, based on the in-degree and out-degree, there only exist two propagation operators $\mathbf{A}_d$ and $\mathbf{A}_d^T$. 
    As for the case of a 2-hop neighborhood, $k$ is set to 6. 
    This is due to the introduction of complex connection patterns arising from directed topology, resulting in four additional propagation operators, as illustrated in Fig.~\ref{fig: model_framework} (i.e., $\left(\mathbf{A}_d\mathbf{A}_d, \mathbf{A}_d^T \mathbf{A}_d^T, \mathbf{A}_d\mathbf{A}_d^T, \mathbf{A}_d^T\mathbf{A}_d\right)$).
    Expanding on this concept, the consideration of an $N$-hop neighborhood implies that $k=2^1+\dots+2^N$.
    While the computational expenses associated with considering higher-order neighborhoods exhibit exponential growth, we can alleviate the reliance on larger values of $k$ by strategically choosing the number of propagation steps. 
    This stems from a pivotal insight obtained during our investigation: \textit{In directed topology, significant connection patterns often emerge in second-order operators, the higher-order operators with specific second-order terms is equal importance}. 
    Consequently, we can opt for a smaller value of $k$ and depend on a larger $K$ to reveal higher-order dependencies for the current node by well-designed propagation.
    It is worth noting that, thanks to training-independent sparse-dense matrix multiplication (i.e., $\mathbf{G}_d$ and $\mathbf{X}$), the feature propagation process experiences a notable acceleration, resulting in a substantial reduction in computation complexity.
    Meanwhile, in reality, feature transformation can be performed with significantly less cost due to better parallelism of sparse matrix multiplications.

    In addition to the aforementioned key insights on reducing computation complexity, we further recommend a careful selection of suitable DP operators using Eq.~(\ref{eq: coefficient}) to improve running efficiency. 
    Specifically, under the assumption of known labels for part of nodes, this equation quantifies the correlation between a specific propagation rule and node profiles.
    Subsequently, we recommend selecting $\mathbf{G}_d$ with a higher value of $r(\mathbf{G}_d, \mathbb{N})$ to construct DPs.
    This is because a larger and positive value of $r(\mathbf{G}_d, \mathbb{N})$ reveals a strong correlation between the current propagation rule $\mathbf{G}_d$ and the distribution of node profiles. 
    This guides fine-grained feature propagation, encompassing diverse connection patterns (i.e., homophily and heterophily). 
    In other words, we do not confine the feature propagation process to homophilous or heterophilous node pairs. 
    Instead, we globally discover systematic connection rules and emphasize this through the DP operator.
    For instance, in Fig.\ref{fig: amud_toy_example}, $\mathbf{A}_d\mathbf{A}_d^T$ and $\mathbf{A}_d^T\mathbf{A}_d$ capture homophily among Node1, Node2, and Node3 under the consideration of directed topology. 
    However, the heterophilous edges revealed by $\mathbf{A}_d\mathbf{A}_d$ and $\mathbf{A}_d^T\mathbf{A}_d^T$ lack clear rules (\sethlcolor{red!0}\hl{Node 7} $\to$ \sethlcolor{blue!0}\hl{Node 1}, \sethlcolor{red!0}\hl{Node 5} $\to$ \sethlcolor{green!0}\hl{Node 4}, and \sethlcolor{blue!0}\hl{Node 1} $\to$ \sethlcolor{green!0}\hl{Node 6}), resulting in relatively small negative values. 
    In contrast, if heterophilous connections exhibit explicit associations, this will manifest as larger positive values (e.g., $\mathbf{A}_d\mathbf{A}_d$ and $\mathbf{A}_d^T\mathbf{A}_d^T$ reflects \sethlcolor{blue!0}\hl{blue} $\to$ \sethlcolor{green!0}\hl{green}).

\subsection{Node-adaptive Attention Mechanisms}
\label{sec: Node-adaptive Attention Mechanisms}

    After $K$-step feature propagation in Eq.~(\ref{eq: k-order DP operator}), we get a list of propagated features under different steps  $\left[\tilde{\mathbf{X}}^{(1)},\dots,\tilde{\mathbf{X}}^{(K)}\right]$, which contains DP-guided multi-scale structural encoding $\tilde{\mathbf{X}}^{(l)}\in\mathbb{R}^{n\times{(k+1)}f}$ and propagation-guided multi-granularity node representations $\tilde{\mathbf{X}}\in\mathbb{R}^{n\times{K(k+1)}f}$.
    These messages comprehensively capture node interaction under the influence of directed topology. 
    However, not all messages contribute to label prediction, especially when considering more granular details at the node level. 
    In other words, due to distinct graph contexts associated with different nodes, the required structural encoding varies for each node.
    To address this issue, we propose the following two hierarchical attention mechanisms to achieve end-to-end node-adaptive representation fusion. 

    \begin{algorithm}[t]
    \caption{Our Proposal's AMUD and ADPA Workflow}
    \label{alg: amud_adpa}
    \begin{algorithmic}[1]
    \STATE \underline{\emph{AMUD\;Topological\;Modeling\;Guidance}}:
        \STATE Generating $k$-order DP operators $\mathbf{G}_d$ from $\mathbf{A}_d$;
        \STATE Quantifying the coefficient of node profiles and directed topology according to the Eq.~(\ref{eq: coefficient});
        \STATE Calculating AMUD score $S$ according to Eq.~(\ref{eq: guidance_score});
        \IF {$S<\theta$} 
            \STATE ${\mathbf{A}}_u=\operatorname{Undirected}\;\operatorname{Transformation}\left({\mathbf{A}}_d\right)$;
        \ELSE
            \STATE ${\mathbf{A}}_d={\mathbf{A}}_d$;
        \ENDIF
    \STATE \underline{\emph{ADPA\;Forward\;Propagation\;Running\;Pipeline\;}}:
        \FOR {each node $i=1,\cdots ,n$}
            \STATE Each node propagates its message according to Eq.~(\ref{eq: k-order DP operator});
            \STATE Each node aggregates $k$-order multi-scale structural encoding messages according to Eq.~(\ref{eq: node-wise dp attention});
            \STATE Each node aggregates $K$-hop multi-granularity node representation messages according to Eq.~(\ref{eq: JK attention});
            \STATE Each node executes forward propagation by $\operatorname{MLP}$ and current fusion representation to predict node label $\hat{\mathbf{Y}}_i$;
        \ENDFOR
    \end{algorithmic}
    \end{algorithm}
    
\noindent
\textbf{Node-wise DP Attention.} 
    Since $k$-order DP operators contribute diverse types of directed information flows to the current node, our objective is to dynamically aggregate the aforementioned multi-scale structural encoding and the initial residual efficiently using the learnable weights $\mathbf{W}_{DP}\in\mathbb{R}^{n\times (k+1)}$ in the node-wise DP attention.
    In other words, in the first level of two hierarchical node-wise attention, we focus on the contributions of different DP operators at each propagation step.
    Formally, the above process is defined as
    
\begin{equation}
    \label{eq: node-wise dp attention}
    \begin{aligned}
        \overline{\mathbf{X}}_i^{(l)} \!\!\!&= \operatorname{MLP}\!\left(\!\mathbf{W}_{DP}^{i,1} \mathbf{X}^{(0)}_i\Vert\mathbf{W}_{DP}^{i,2}\mathbf{X}_{i,\mathbf{G}_{d_1}}^{(l)}\!\!\left\Vert\cdots \right\Vert\!\mathbf{W}_{DP}^{i,k+1} \mathbf{X}_{i,\mathbf{G}_{d_k}}^{(l)}\!\right),\\
    \end{aligned}
\end{equation}
    where $\operatorname{MLP}:=\mathbb{R}^{n\times(k+1)f}\rightarrow\mathbb{R}^{n\times f}$ represents the node-adaptive multi-scale representation fusion function and $\cdot||\cdot$ denotes the feature concatenation.
    Now, we have obtained $\overline{\mathbf{X}}^{(l)}\in\mathbb{R}^{n\times f}$, representing the adaptively acquired node representation at each propagation step. 
    Subsequently, considering a $K$-step propagation process, we obtain the corresponding list $\tilde{\mathbf{X}}=\left[\overline{\mathbf{X}}^{(1)},\dots,\overline{\mathbf{X}}^{(K)}\right]$ of length $K$. 
    It is worth noting that the attention mechanism utilized in Eq.~(\ref{eq: node-wise dp attention}) can be substituted with any other reasonable attention mechanisms for altering the methods of feature aggregation in the $\operatorname{MLP}$, such as Gate attention~\cite{ahmad2021gate}, recursive attention~\cite{gamlp}, and JK attention~\cite{xu2018jknet}.
    To further explore the impact of the aforementioned different attention mechanisms in our implementation, we conduct a series of ablation experiments in Sec.~\ref{sec: Ablation Study and Sensitivity Analysis}.

\noindent
\textbf{Node-wise Hop Attention.} 
    As we all know, in the graph learning process based on message-passing, each node gradually receives information from its low-order and high-order neighbors. 
    We refer to this collection of neighborhood nodes as the receptive field of the current node. 
    Recent research~\cite{frasca2020sign,sun2021sagn,zhang2021rod,2021fsgnn,lee2023aerognn} have emphasized that the most appropriate receptive field for each node varies due to the intricate contextual information inherent within the graph.
    Notably, this is particularly crucial for digraphs, where high-order topological structure often entails valuable cues for predictions. 
    Moreover, distinct differences exist among various high-order structures, as analyzed in Sec.~\ref{sec: AMUD: Graph-based Data Engineering}.
    Inspired by these key insights and empirical studies, we advocate explicitly learning the importance and relevance of multi-granularity knowledge in a node-adaptive manner. 
    To this end, we introduce the second level of two hierarchical node-adaptive attention, which automatically leverages knowledge from different neighborhoods guided by various DP to boost performance.
\begin{equation}
    \label{eq: JK attention}
    \begin{aligned}
        &{\mathbf{X}}_i^\star=\sum_{l=1}^{K} \mathbf{W}_{hop}^{(l)} \mathbf{\overline{X}}_i^{(l)},\;\mathbf{W}_{hop}^{(l)} = e^{\delta(\mathbf{E}_i^{(l)})} / \sum_{k=1}^{K}e^{\delta(\mathbf{E}_i^{(k)})},\\
        &\;\;\;\;\;\mathbf{E}_i^{(l)} = \operatorname{MLP}\left(\mathbf{\overline{X}}_i^{(1)}||\mathbf{\overline{X}}_i^{(2)}||\cdots||\mathbf{\overline{X}}_i^{(K-1)}||\mathbf{\overline{X}}_i^{(K)}\right),\\
    \end{aligned}
\end{equation}
    where $\delta$ is the non-linear activate function, $\mathbf{W}_{hop}^{(l)}$ and $\mathbf{E}_i^{(l)}$ are used for computing the node-wise attention weights. 
    This hop attention mechanism is designed to construct a personalized multi-granularity representation fusion for each node, facilitating the learning of message aggregation weights by the attention mechanism. 
    These learned weights are subsequently input into the attention-based combination branch, generating a refined attention feature representation for each node. 
    As the training progresses, the attention-based combination branch gradually accentuates the importance of neighborhood regions that contribute more significantly to the target nodes.

\subsection{Complexity Analysis}
\label{sec: complexity analysis}
    In this section, we present an algorithm complexity analysis of ADPA to illustrate how the decoupled model architecture provides users with a computation-friendly paradigm.
    Specifically, let $n,m$, and $f$ be the number of nodes, edges, and feature dimensions. 
    $k$ and $K$ correspond to the number of $k$-order DP operators and the propagation steps. 
    $L$ refers to the number of layers in the $\operatorname{MLP}$ classifier. 
    Since ADPA utilizes $k$-order DP in the feature propagation, the overall time complexity is $\mathcal{O}(kKmf)$. 
    Notably, this process is independent of the training phase and can be pre-processed and cached in local memory.
    As a result, the time and space complexity during the training is negligible.
    Furthermore, the time complexity of $\operatorname{MLP}$ training is $\mathcal{O}(kLnf^2)$, due to the fact that each feature generated from $k$-order DP operators is fed into the $\operatorname{MLP}$ of layer $L$. 
    The space complexity of ADPA is composed of features in the feature propagation and the neural network during training. 
    Thus, the overall space complexity of ADPA is $\mathcal{O}(kf+kLf^2)$.
    Notably, feature transformation $\mathcal{O}(Lnf^2)$ can be performed with significantly less cost due to better parallelism of dense-dense matrix multiplications.

\subsection{A Unified Paradigm for Spatial-based Directed GNNs}
\label{sec: Comparison between the Spatial-based GNNs and ADPA}

    In a nutshell, ADPA comprises two pivotal modules: directed patterns guided feature propagation (see Sec.~\ref{sec: directed patterns guided feature propagation}) and two hierarchical node-adaptive attention mechanisms (see Sec.~\ref{sec: Node-adaptive Attention Mechanisms}), which constitute a general paradigm for spatial-based directed GNNs.  
    Specifically, we propose that adhering to the spatial domain message-passing digraph learning paradigm should involve the following two steps:
    \textbf{Step 1}: Define appropriate DP operators to perform fine-grained feature propagation, uncovering genuine patterns within intricate topologies.
    \textbf{Step 2}: Achieve end-to-end training through carefully designed message aggregation functions.
    Building upon this conception, we will elaborate on the comparison between ADPA and existing spatial-based directed GNNs.

\noindent
    (1) 
    DGCN~\cite{tong2020dgcn} and Dir-GNN~\cite{dirgnn_rossi_2023} limit their consideration to incomplete 2-order DPs and employ outdated learnable message aggregation mechanisms, which can be regarded as a special case of hop attention.
    Notably, Sec.~\ref{sec: AMUD: Graph-based Data Engineering} demonstrates that other DPs also provide valuable structural insights.
    In light of this, ADPA extends DP operators when propagating node features and achieves multi-granularity message fusion by two hierarchical node-adaptive attention mechanisms.

\noindent
    (2) 
    A2DUG~\cite{maekawa2023a2dug} solely focuses on the undirected versions of propagation operators, obscuring the homophily and heterophily inherent in directed edges. 
    As discussed in Sec.~\ref{sec: directed modeling guidance}, there exist significant semantic distinctions among DPs of the same order. 
    Converting directed edges into undirected ones also overlooks such crucial topological information. 
    Consequently, ADPA incorporates various directed graph patterns of the same orders, thereby enhancing the model's performance.

\noindent
    (3) 
    NSTE~\cite{kollias2022nste} and DIMPA~\cite{he2022dimpa} adhere to a tightly coupled model architecture, utilizing two sets of weights to encode the in- and out-degree edges. 
    Simultaneously, they expand the receptive field between model layers, resulting in unacceptable recursive computation costs. 
    This can be viewed as a combination of first-order DP and sub-optimal attention constrained by in- and out-degree. 
    In contrast, ADPA follows a decoupled design shown in Sec.~\ref{sec: Architecture Overview} principle, maximizing computational efficiency while enjoying superior predictive performance.

\section{Experiments}
    In this section, we present a comprehensive evaluation of our proposed graph-based data engineering framework AMUD and digraph learning paradigm ADPA. 
    We first introduce 16 graph benchmark datasets commonly used in graph learning, including homophily and heterophily which have been generally recognized by previous studies. 
    We then offer detailed descriptions of baselines, including state-of-the-art directed GNNs and undirected GNNs designed for homophily and heterophily. 
    We also present detailed settings for replicating our experimental results.
    After that, we aim to address the following questions:
    \textbf{Q1}: Does the modeling guidance of directed topology provided by AMUD prove to be effective?
    \textbf{Q2}: Can ADPA achieve {better} predictive performance than state-of-the-art baselines under both undirected and directed scenarios?
    \textbf{Q3}: If ADPA is effective, what contributes to its performance gain?
    \textbf{Q4}: How does ADPA perform under the sparse settings for digraphs, such as low label/edge rate and missing features?

\begin{table*}[t]
\caption{{The statistical information of the experimental datasets, E.Homo and Adj.Homo are edge and adjusted homophily in Sec.~\ref{sec: Undirected Graph Neural Networks}.}
}
\footnotesize 
\label{tab: datasets}
\resizebox{\linewidth}{30mm}{
\setlength{\tabcolsep}{1.5mm}{
\begin{tabular}{c|ccccccccc}
\midrule[0.3pt]
Datasets         & \#Nodes & \#Edges & \#Features & \#Classes & \#Train/Val/Test & \#E.Homo & \#Adj.Homo & \#AMUD-Score & Description            \\ \midrule[0.3pt]
CoraML           & 2,995   & 8,416   & 2,879      & 7         & 140/500/2,355    & 0.792    & 0.784      & \textcolor{black}{0.380}(U-)         & citation network       \\
CiteSeer         & 3,312   & 4,715   & 3,703      & 6         & 120/500/2,692    & 0.739    & 0.726      & \textcolor{black}{0.269}(U-)          & citation network       \\
PubMed           & 19,717  & 88,648  & 500        & 3         & 60/500/1,000     & 0.802    & 0.782      & -            & citation network       \\
Tolokers         & 11,758  & 519,000 & 10         & 2         & 50\%/25\%/25\%   & 0.595    & 0.530      & \textcolor{black}{0.405}(U-)         & crowd-sourcing network  \\
WikiCS           & 11,701  & 290,519 & 300        & 10        & 580/1769/5847    & 0.689    & 0.674      & \textcolor{black}{0.392}(U-)         & web-link network        \\
Amazon-computers & 13,752  & 287,209 & 767        & 10        & 200/300/12,881   & 0.786    & 0.769      & \textcolor{black}{0.314}(U-)         & co-purchase network    \\ 
ogbn-arxiv       & 169,343 & 2,315,598  &128      & 40        & 91k/30k/48k      & 0.655	& 0.641      & \textcolor{black}{0.469}(U-)           & citation network    \\
Genius          & 421,961	&984,979  &12	    &2	         &50\%/25\%/25\%     & 0.618	&0.558	& \textcolor{black}{0.705}(D-)         & social network         \\
\midrule[0.3pt]
Texas            & 183     & 279     & 1,703      & 5         & 48\%/32\%/20\%   & 0.061    & -0.014     & \textcolor{black}{0.814}(D-)        & web-page network        \\
Cornell          & 183     & 298     & 1,703      & 5         & 48\%/32\%/20\%   & 0.122    & 0.026      & \textcolor{black}{0.712}(D-)        & web-page network        \\
Wisconsin          & 251     & 450     & 1,703      & 5         & 48\%/32\%/20\%   & 0.178    & 0.110      & \textcolor{black}{0.685}(D-)        & web-page network        \\
Chameleon        & 890     & 13,584  & 2,325      & 5         & 48\%/32\%/20\%   & 0.245    & 0.203      & \textcolor{black}{0.657}(D-)        & wiki-page network      \\
Squirrel         & 2,223   & 65,718  & 2,089      & 5         & 48\%/32\%/20\%   & 0.216    & 0.173      & \textcolor{black}{0.693}(D-)        & wiki-page network      \\
Actor            & 7,600   & 26,659  & 932        & 5         & 48\%/32\%/20\%   & 0.217    & 0.172      & \textcolor{black}{0.356}(U-)         & actor network          \\
Roman-empire     & 22,662  & 32,927  & 300        & 18        & 50\%/25\%/25\%   & 0.047    & 0.025      & \textcolor{black}{0.642}(D-)        & article syntax network \\
Amazon-rating    & 24,492  & 93,050  & 300        & 5         & 50\%/25\%/25\%   & 0.380    & 0.334      & \textcolor{black}{0.395}(U-)         & rating network         \\

\midrule[0.3pt]
\end{tabular}
}}
\vspace{-0.2cm}
\end{table*}

\subsection{Experimental Setup}
\label{sec: experimental setup}
{
\noindent \textbf{Datasets.}
    In this section, we evaluate the performance of AMUD and ADPA on 16 digraph/graph benchmark datasets, considering both homophily and heterophily, as generally acknowledged by previous studies~\cite{2020h2gcn,platonov2023hete_gnn_survey4,2021linkx}.
    The majority of these datasets consist of natural directed graphs, with the exception of PubMed.
    For homophily, we perform experiments on 4 citation networks (CoraML, CiteSeer, PubMed, ogbn-arxiv)~\cite{bojchevski2018coraml_citeseer, Yang16cora, hu2020ogb}, crowd-sourcing dataset (Toloklers), web-link dataset (WikiCS)~\cite{shchur2018amazon_datasets}, and co-purchase dataset (Amazon-computers)~\cite{shchur2018amazon_datasets}. 
    Regarding heterophily, we conduct experiments on three web-page networks (Texas, Cornell, and Wisconsin from the WebKB datasets)~\cite{pei2020geomgcn}, two updated wiki-page networks (Chameleon and Squirrel)~\cite{platonov2023hete_gnn_survey4}, movie network (Actor)~\cite{pei2020geomgcn}, syntax network (Roman-empire)~\cite{platonov2023hete_gnn_survey4}, e-commerce network (Amazon-rating)~\cite{platonov2023hete_gnn_survey4}, and social network (Genius)~\cite{2021linkx}. 
    For more statistical information, please refer to Table~\ref{tab: datasets}. Since AMUD is tailored for directed scenarios, there is no need to report the score for naturally undirected PubMed.}

{\noindent \textbf{Baselines.}
    To achieve a comprehensive comparison, we employ
    (1) undirected spatial GCN~\cite{kipf2016gcn},GCNII~\cite{chen2020gcnii}, LINKX~\cite{2021linkx}, GloGNN~\cite{2022glognn}, and AEROGNN~\cite{lee2023aerognn};
    (2) undirected spectral SGC~\cite{wu2019sgc}, GRAND~\cite{chamberlain2021grand}, GPR-GNN~\cite{chien2021gprgnn}, BerNet~\cite{he2021bernnet}, and JacobiConv~\cite{pmlr2022Jacobigcn};
    (3) directed spatial DGCN~\cite{tong2020dgcn}, NSTE~\cite{kollias2022nste}, DIMPA~\cite{he2022dimpa}, DirGNN~\cite{dirgnn_rossi_2023}, and A2DUG~\cite{maekawa2023a2dug}.
    (4) directed spectral DiGCN~\cite{tong2020digcn} and MagNet~\cite{zhang2021magnet}.
    For dataset split, we are aligned with previous studies~\cite{bojchevski2018coraml_citeseer,zhang2021magnet,platonov2022hete_gnn_survey5}.}
    In order to alleviate the influence of randomness, we repeat each experiment 10 times to represent the unbiased performance.
    To ensure fairness, we report the performance of feeding the undirected transformation (U-) of digraphs into the undirected GNNs. 
    Regarding directed GNNs, we default to reporting prediction accuracy on directed inputs (D-).
    Given the numerous baselines, we strive to diversify their usage in subsequent experiments, ensuring comprehensive comparisons without complex charts and improving the readability of the results.

\noindent \textbf{Hyper-parameters.}
    The hyperparameters are set based on the original paper if available.
    Otherwise, we perform an automatic hyperparameter search via the Optuna~\cite{akiba2019optuna}.
    For our proposed ADPA, the steps of feature propagation and MLP layers are explored within the ranges of 1 to 5.
    We set the hidden dimension to 64 and explore the optimal convolution kernel coefficient within the ranges of 0 to 1.

\noindent \textbf{Environment.}
    To ensure reproducibility, we provide the hardware settings, including a machine with Intel(R) Xeon(R) Gold 6230R CPU \@ 2.10GHz, and NVIDIA GeForce RTX 3090 with 24GB memory and CUDA 11.8.
    The operating system is Ubuntu 18.04.6 with 216GB memory.

\begin{table}[t]
\setlength{\abovecaptionskip}{0.1cm}
\setlength{\belowcaptionskip}{0.1cm}
\caption{{Performance in homophilous ($\operatorname{Score}<0.5$) datasets.}
}
\footnotesize
\label{tab: homo_cmp}
\resizebox{\linewidth}{30mm}{
\setlength{\tabcolsep}{1.2mm}{
\begin{tabular}{c|ccccccc}
\midrule[0.3pt]
Model         & CoraML   & CiteSeer & PubMed   & Tolokers & WikiCS   & \begin{tabular}[c]{@{}c@{}}Amazon\\ Computers\end{tabular} & Rank     \\ \midrule[0.3pt]
GCN           & 84.2±0.5 & 65.3±0.5 & 79.2±0.4 & 79.0±0.5 & 77.5±0.3 & 78.4±0.8                                                   & 6.8          \\
SGC           & 83.8±0.2 & 62.8±0.3 & 78.6±0.2 & 78.8±0.2 & 76.4±0.2 & 77.3±0.4                                                   & 9.7          \\
GCNII         & 84.5±0.6 & 65.1±0.4 & 79.4±0.3 & 79.2±0.4 & 77.8±0.3 & 78.6±0.7                                                   & 5.8          \\
GRAND         & 83.9±0.4 & 63.8±0.5 & 79.0±0.4 & 78.7±0.3 & 77.2±0.4 & 78.5±0.6                                                   & 7.9          \\
LINKX         & 83.9±0.3 & 63.3±0.4 & 77.6±0.5 & 78.0±0.6 & 77.2±0.4 & 77.9±0.7                                                   & 9.2          \\
BerNet        & 83.4±0.4 & 64.5±0.6 & 78.4±0.5 & 77.6±0.4 & 77.4±0.8 & 80.1±0.9                                                   & 5.4          \\
JacobiConv    & 84.1±0.7 & 64.8±0.9 & 79.6±0.3 & 78.3±0.4 & 78.0±0.5 & 79.3±1.0                                                   & 3.7          \\
GPRGNN        & 84.7±0.7 & 64.7±0.7 & 79.3±0.4 & 79.2±0.3 & 77.4±0.6 & 78.5±1.2                                                   & 8.3          \\
GloGNN        & 84.4±0.8 & 63.5±1.0 & 78.8±0.6 & 78.5±0.5 & 78.3±0.2 & 79.7±1.0                                                   & 5.4          \\
AERO-GNN      & 83.9±0.8 & 63.0±0.6 & 79.5±0.6 & 78.9±0.4 & 78.0±0.4 & 79.5±0.9                                                   & 3.6          \\ \midrule[0.3pt]
DGCN          & 83.1±1.0 & 63.2±0.8 & 77.9±0.4 & 77.8±0.4 & 76.6±0.4 & 77.4±0.4                                                   & 14.2           \\
DiGCN         & 83.5±0.6 & 64.7±0.6 & 78.4±0.4 & 78.3±0.3 & 77.2±0.3 & 78.1±0.7                                                   & 12.4         \\
MagNet        & 84.4±0.6 & 63.8±0.4 & 78.6±0.6 & 77.8±0.2 & 76.0±0.6 & 78.3±0.3                                                   & 7.8          \\
NSTE          & 84.0±0.3 & 64.9±0.7 & 77.8±0.7 & 77.4±0.4 & 76.2±0.5 & 77.8±0.8                                                   & 10.3         \\
DIMPA         & 83.6±0.5 & 64.8±0.9 & 78.0±0.4 & 78.8±0.2 & 76.3±0.3 & 78.0±0.5                                                   & 10.7         \\
DirGNN        & 83.8±0.9 & 64.2±0.6 & 78.5±0.9 & 78.6±0.3 & 77.5±0.4 & 78.3±0.4                                                   & 13.7           \\
A2DUG         & 83.7±1.0 & 64.8±0.8 & 78.8±0.8 & 78.0±0.9 & 76.8±0.7 & 77.5±0.8                                                   & 12.3         \\
\cellcolor[HTML]{d0cfcf}{ADPA} & \cellcolor[HTML]{d0cfcf}84.5±0.6 & \cellcolor[HTML]{d0cfcf}66.0±0.4 & \cellcolor[HTML]{d0cfcf}80.2±0.4 & \cellcolor[HTML]{d0cfcf}80.7±0.4 & \cellcolor[HTML]{d0cfcf}79.4±0.5 & \cellcolor[HTML]{d0cfcf}80.9±0.4                                                   & \cellcolor[HTML]{d0cfcf}{1.2} \\ \midrule[0.3pt]
\end{tabular}
}}

\end{table}

\begin{table}[t]
\setlength{\abovecaptionskip}{0.1cm}
\setlength{\belowcaptionskip}{0.1cm}
\caption{{Performance in heterophilous ($\operatorname{Score}>0.5$) datasets.}
}
\footnotesize
\label{tab: hetero_cmp}
\resizebox{\linewidth}{30mm}{
\setlength{\tabcolsep}{1.2mm}{
\begin{tabular}{c|ccccccc}
\midrule[0.3pt]
Model         & Texas    & Cornell  & Wisconsin & Chameleon & Squirrel & \begin{tabular}[c]{@{}c@{}}Roman\\ Empire\end{tabular} & Rank       \\ \midrule[0.3pt]
GCN           & 69.6±3.0 & 60.3±4.2 & 65.6±2.8  & 41.9±1.1  & 34.6±0.9 & 78.9±0.3                                               & 13.2         \\
SGC           & 64.3±1.2 & 55.6±1.0 & 54.5±0.8  & 36.9±0.1  & 38.1±0.1 & 54.2±0.1                                               & 15.9       \\
GCNII         & 69.2±3.1 & 60.9±3.9 & 64.9±3.4  & 42.2±0.9  & 34.2±0.8 & 79.4±0.2                                               & 13         \\
GRAND         & 65.8±2.8 & 58.4±3.8 & 65.2±2.9  & 40.7±1.6  & 33.1±1.3 & 77.8±0.4                                               & 14.7         \\
LINKX         & 77.0±2.9 & 76.8±5.2 & 73.7±4.1  & 38.9±2.3  & 39.8±1.2 & 77.5±0.6                                               & 13.3         \\
BerNet        & 78.4±2.4 & 78.5±2.3 & 78.6±1.9  & 38.6±1.8  & 37.9±0.8 & 77.8±0.5                                               & 10.7       \\
JacobiConv    & 76.8±3.6 & 79.0±4.4 & 77.5±2.6  & 40.2±2.4  & 37.4±0.9 & 76.3±0.2                                               & 12.3         \\
GPRGNN        & 77.5±3.4 & 77.9±3.1 & 76.3±3.9  & 41.3±1.5  & 37.0±0.9 & 78.5±0.4                                               & 11.7       \\
GloGNN        & 82.2±3.1 & 79.6±3.0 & 78.0±2.2  & 40.2±2.6  & 41.3±1.2 & 78.2±0.3                                               & 7.7        \\
AERO-GNN      & 80.4±2.7 & 80.4±2.8 & 78.6±3.6  & 41.0±2.9  & 41.8±1.1 & 76.8±0.3                                               & 7        \\ \midrule[0.3pt]
DGCN          & 75.8±3.5 & 77.9±3.1 & 76.9±1.4  & 42.3±2.3  & 40.6±1.0 & 80.0±0.7                                               & 10.3       \\
DiGCN         & 79.5±3.2 & 77.8±4.9 & 77.2±2.2  & 43.4±1.8  & 42.0±1.7 & 81.3±0.4                                               & 7.9        \\
MagNet        & 80.5±2.1 & 79.4±3.5 & 78.4±2.6  & 44.5±1.1  & 42.7±1.5 & 81.9±0.3                                               & 3.7        \\
NSTE          & 78.4±3.0 & 78.6±2.3 & 77.6±3.8  & 42.2±2.6  & 41.9±0.9 & 81.2±0.4                                               & 5.3        \\
DIMPA         & 79.5±2.0 & 79.3±3.1 & 78.4±3.1  & 43.8±1.6  & 41.2±1.0 & 79.8±0.2                                               & 4.7        \\
DirGNN        & 81.4±2.4 & 80.0±3.6 & 79.6±3.8  & 44.6±1.7  & 42.5±0.8 & 81.8±0.3                                               & 3.9          \\
A2DUG         & 80.5±3.9 & 80.5±4.3 & 78.2±4.6  & 43.8±2.8  & 42.8±1.1 & 81.3±0.5                                               & 6.3        \\
\cellcolor[HTML]{d0cfcf}{ADPA} & \cellcolor[HTML]{d0cfcf}83.8±2.7 & \cellcolor[HTML]{d0cfcf}82.9±3.0 & \cellcolor[HTML]{d0cfcf}81.6±3.5  & \cellcolor[HTML]{d0cfcf}46.2±1.3  & \cellcolor[HTML]{d0cfcf}45.2±1.3 & \cellcolor[HTML]{d0cfcf}84.3±0.3                                               & \cellcolor[HTML]{d0cfcf}{1} \\ \midrule[0.3pt]
\end{tabular}
}}
\end{table}

\subsection{AMUD Guidance}
\label{sec: AMUD Guidance}
    {To answer \textbf{Q1}, we report the experimental results on AMUndirected ($\operatorname{Score}<0.5$) and AMDirected ($\operatorname{Score}>0.5$) in Table~\ref{tab: homo_cmp} and Table~\ref{tab: hetero_cmp}, where Rank represents the average ranking of prediction accuracy.
    The statistics of these datasets adhere to widely recognized conclusions measuring topological properties from prior studies (homophilous measures) and the key insights mentioned in Sec.~\ref{sec: introduction} and Sec.~\ref{sec: AMUD: Graph-based Data Engineering}: \textit{Intuitively, undirected and directed topological modeling are respectively applicable to homophily and heterophily as reflected in edge and adjusted homophily shown in Table~\ref{tab: datasets}}.
    However, AMUD identifies three abnormal cases: Genius, Actor, and Amazon-rating. 
    Despite their being considered representative benchmark datasets, AMUD provides starkly opposite modeling guidance by quantifying the correlation between directed topology and nodes.
    To emphasize this phenomenon, we separately report the experimental results in Table~\ref{tab: anomaly_cmp} with a thorough analysis.}

\begin{table}[t]
\setlength{\abovecaptionskip}{0.3cm}
\setlength{\belowcaptionskip}{0cm}
\caption{{Improvement from the AMUD (U- or D-).}
}
\vspace{-0.2cm}
\footnotesize
\label{tab: anomaly_cmp}
\resizebox{\linewidth}{45mm}{
\setlength{\tabcolsep}{0.7mm}{
\begin{tabular}{ccccccc}
\midrule[0.3pt]
Model-1 & Model-2  & Actor-1  & Actor-2   & Rating-1 & Rating-2 & $\Rightarrow$ \\ \midrule[0.3pt]
\midrule[0.3pt]
GCN     & Jacobi   & 37.7±0.5 & 37.8±0.4  & 46.8±0.4 & 47.1±0.5 & Undir        \\
LINKX   & GloGNN   & 37.4±0.5 & 37.9±0.8  & 46.7±0.3 & 46.7±0.5 & Undir        \\
BerNet  & AERO     & 38.1±0.7 & 38.2±0.6  & 46.5±0.4 & 46.9±0.6 & Undir        \\ \midrule[0.3pt]
\midrule[0.3pt]
Model   & U-A      & D-A      & $\Uparrow$ & U-R      & D-R      & $\Uparrow$    \\ \midrule[0.3pt]
\midrule[0.3pt]
MagNet  & \textbf{37.2±0.9} & 35.5±0.8 & 4.8\%     & \textbf{46.5±0.5} & 44.4±0.5 & 4.3\%        \\
DIMPA   & \textbf{38.0±0.4} & 36.3±0.8 & 3.5\%     & \textbf{46.3±0.4} & 44.7±0.5 & 4.5\%        \\
DirGNN  & \textbf{38.2±0.3} & 35.9±0.4 & 5.6\%     & \textbf{46.5±0.5} & 44.1±0.4 & 6.7\%        \\
ADPA    & \textbf{39.7±0.7} & 38.8±0.3 & 1.7\%     & \textbf{49.0±0.3} & 48.2±0.4 & 2.3\%        \\ \midrule[0.3pt]
\midrule[0.3pt]
Model-1 & Model-2  & arxiv-1  & arxiv-2   & Genius-1 & Genius-2 & $\Rightarrow$ \\ \midrule[0.3pt]
\midrule[0.3pt]
GCN     & Jacobi   & 71.9±0.2 & 72.6±0.3  & 88.4±0.4 & 89.6±0.5 & Undir        \\
LINKX   & GloGNN   & 71.5±0.4 & 72.5±0.3  & 89.2±0.3 & 89.2±0.4 & Undir        \\
BerNet  & AERO     & 72.0±0.3 & 72.4±0.4  & 88.9±0.3 & 89.1±0.6 & Undir        \\ \midrule[0.3pt]
\midrule[0.3pt]
Model   & U-a      & D-a      & $\Uparrow$ & U-G      & D-G      & $\Uparrow$    \\ \midrule[0.3pt]
\midrule[0.3pt]
MagNet  & \textbf{72.7±0.3} & 70.4±0.3 & 3.3\%     & 88.7±0.5 & \textbf{91.8±0.4} & 3.5\%        \\
DIMPA   & \textbf{73.2±0.2} & 70.0±0.3 & 4.2\%     & 88.4±0.5 & \textbf{91.3±0.3} & 3.4\%        \\
DirGNN  & \textbf{73.0±0.3} & 70.5±0.4 & 3.5\%     & 89.2±0.6 & \textbf{92.0±0.5} & 3.1\%        \\
ADPA    & \textbf{73.8±0.3} & 73.0±0.3 & 1.2\%     & 91.7±0.4 & \textbf{92.8±0.3} & 1.3\%        \\ \midrule[0.3pt]
\end{tabular}
}}

\vspace{-0.4cm}
\end{table}
    {\underline{\textit{Perspective 1}}.
    To begin with, we validate the effectiveness of the proposed workflow shown in Fig.~\ref{fig: motivation_AMUD_AdpA}: \textit{For AMUndirected, although it can be handled by the directed GNN, we recommend utilizing existing reasonable undirected GNN. 
    As for AMDirected, they can only be fed into directed GNNs.}
    As the average Rank list depicted in Fig.~\ref{fig: motivation_AMUD_AdpA}, we observe that undirected GNNs tend to outperform directed GNNs when $\operatorname{Score}<0.5$, whereas directed GNNs generally excel over undirected GNNs when $\operatorname{Score}>0.5$.
    For instance, in Table~\ref{tab: homo_cmp}, the Rank of undirected BerNet is 5.4 is much higher than directed Dir-GNN (Rank 13.7). 
    As presented in Table~\ref{tab: hetero_cmp}, the rank of BerNet (10.7) is much lower than Dir-GNN (3.9) in dealing with heterophilous dataset, which also validates the effectiveness of our mechanism.
    Furthermore, the average Rank of directed and undirected methods also substantiates this point. 
    Meanwhile, although ADPA does not consistently achieve the best performance in Table~\ref{tab: homo_cmp}, we observe it to be a competitive method. 
    This supports our claims in Sec.~\ref{sec: preliminaries}: ADPA is a feasible choice for both AMUndirected and AMDirected.}

    {\underline{\textit{Perspective 2}}.
    After demonstrating the validity of our proposal in Fig.~\ref{fig: motivation_AMUD_AdpA}, we further investigate whether AMUD can break the entanglement of homophily and heterophily by examining four specific datasets.
    During our investigation of applying AMUD, we observe that not all heterophilous (homophilous) datasets (measured by the edge and adjusted homophily) can achieve high (low) AMUD scores (threshold is $0.5$). 
    Specifically, Actor and Amazon-rating (Genius) are conventionally regarded as heterophilous (homophilous) datasets. 
    Nevertheless, the AMUD scores of Actor and Amazon-rating (Genius) shown in Table~\ref{tab: datasets} indicate that they should be modeled as undirected (directed) graphs.
    Furthermore, we consider the ogbn-arxiv dataset for comprehensive experimental results.
    Building upon this, we present the pertinent experimental results in Table~\ref{tab: anomaly_cmp} to substantiate our claims.
    We attribute this to AMUD's success in identifying the correlation between directed edges and node profiles in these four datasets. 
    Remarkably, compared to other directed baselines, performing undirected transformations on the input does not significantly impact ADPA. 
    This indirectly validates the robustness of ADPA in facing both undirected and directed scenarios; it is highly adaptive to numerous deployment needs while maintaining excellent performance.}


\begin{figure*}[t]
	\centering
    \setlength{\abovecaptionskip}{0.2cm}
    \setlength{\belowcaptionskip}{-0.2cm}
  \includegraphics[width=\textwidth]{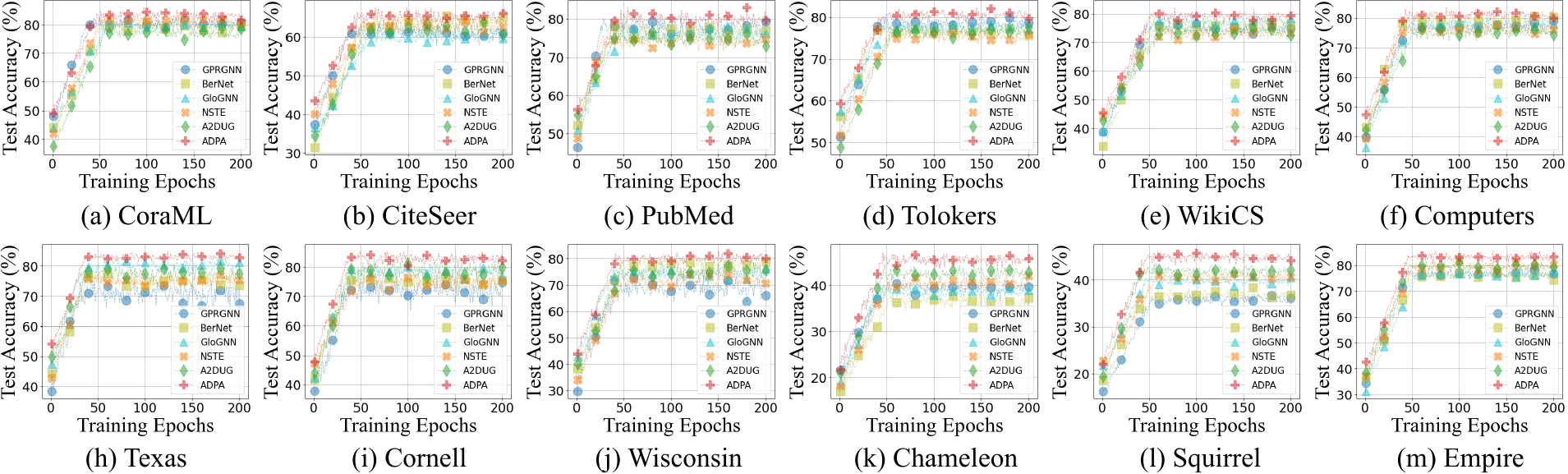}
  \caption{
    Convergence curves on the AMUndirected (upper) and AMDirected (lower).
}
    \vspace{-0.1cm}
  \label{fig:exp_converge}
\end{figure*}

\begin{figure*}[t]
	\centering
    \setlength{\abovecaptionskip}{0.2cm}
    \setlength{\belowcaptionskip}{-0.2cm}
  \includegraphics[width=\textwidth]{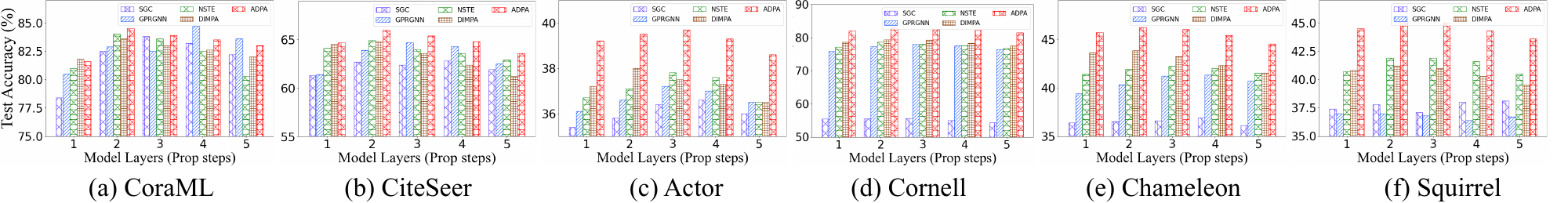}
  \caption{
    AMUndirected (Left three) and AMDirected (Right three) performance under different propagation steps $K$.
}
    \vspace{-0.1cm}
  \label{fig: exp_prop}
\end{figure*}

\subsection{Performance Comparison}
\label{sec: performance comparison}
    To answer \textbf{Q2}, we note that ADPA consistently exhibits superior performance compared to all baselines in Table~\ref{tab: homo_cmp} and Table~\ref{tab: hetero_cmp}.
    For instance, ADPA exhibits a significant lead over the state-of-the-art MagNet on the Roman-empire ($\operatorname{Score}=0.642$) by around 2.9\%. 
    Moreover, ADPA achieves an improvement of approximately 1.5\% compared to the highly competitive JacobiConv on the Citeseer ($\operatorname{Score}=0.269$), even though undirected transformation has been applied during training. 
    These consistent results confirm the superiority of ADPA over other state-of-the-art baselines on intricate topology, attributable to the two pivotal modules: directed patterns guided feature propagation and two hierarchical node-adaptive attention mechanisms. 
    By adaptively aggregating multi-scale messages from various $k$-order neighbors, ADPA achieves effective multi-granularity representation fusion, which results in remarkable improvements in predictive performance.
    
    Building upon the above results, we visualize the training process in Fig.~\ref{fig:exp_converge}. 
    Experimental results demonstrate that ADPA consistently outperforms other methods throughout the entire training process and exhibits more stable convergence. 
    For instance, in the Toloker and WikiCS datasets with $\operatorname{Score}<0.5$, ADPA reaches close-to-optimal performance around the 50th epoch, and the training curves indicate stability in subsequent convergence. 
    Similarly, the same conclusion applies to the Empire dataset with $\operatorname{Score}>0.5$, highlighting the superiority of ADPA.
    Notably, due to the relatively small size of the Texas, Cornell, and Wisconsin datasets, they consistently result in higher variance in unbiased performance reports in Table~\ref{tab: hetero_cmp}. 
    This is also reflected in the visualization results, where their convergence curves often exhibit drastic fluctuations. 
    Consequently, these datasets lead to significant performance variations in models that cannot maintain stability during training, making them challenging to deploy in real-world scenarios, such as GPRGNN and NSTE. 
    In contrast, ADPA demonstrates satisfactory robustness.

\begin{table}[t]
\setlength{\abovecaptionskip}{0.3cm}
\setlength{\belowcaptionskip}{0.2cm}
\caption{ADPA performance under different $k$-order DP operators.
}

\footnotesize
\label{tab: ablation_k-orders}
\resizebox{\linewidth}{24.5mm}{
\setlength{\tabcolsep}{1mm}{
\begin{tabular}{c|ccccc}
\midrule[0.3pt]
Dataset   & 1-order    & 2-order             & 3-order             & 4-order    & 5-order    \\ \midrule[0.3pt]
CoraML    & 79.63±0.48 & \textbf{84.52±0.64} & 82.94±0.54          & 82.17±0.51 & 80.81±0.70 \\
CiteSeer  & 62.72±0.45 & \textbf{66.03±0.38} & 64.11±0.25          & 63.57±0.52 & 63.89±0.46 \\
Actor     & 37.34±0.50 & 38.58±0.64          & \textbf{39.65±0.72} & 38.53±0.66 & 37.94±0.78 \\
Tolokers  & 80.22±0.16 & \textbf{80.72±0.37} & 78.60±0.32          & 78.92±0.42 & 77.68±0.49 \\
Amazon Rating    & 45.47±0.99 & 48.42±0.45          & \textbf{48.96±0.30} & 47.84±0.56 & 48.14±0.49 \\
Computers & 79.68±0.19 & \textbf{80.94±0.42} & 79.25±0.60          & 78.51±0.44 & 77.19±0.85 \\ \midrule[0.3pt]
Texas     & 81.62±2.23 & \textbf{83.78±2.71} & 82.54±2.38          & 81.06±2.53 & 80.54±2.96 \\
Cornell   & 74.68±3.28 & \textbf{82.92±3.04} & 76.93±2.40          & 79.45±2.51 & 78.32±1.72 \\
Wisconsin & 78.04±2.56 & \textbf{81.57±3.51} & 75.49±4.47          & 74.51±4.16 & 74.12±5.08 \\
Chameleon & 44.02±1.77 & \textbf{46.19±1.34} & 41.96±0.59          & 43.24±0.73 & 44.03±1.75 \\
Squirrel  & 41.11±1.36 & \textbf{45.22±1.28} & 43.96±1.22          & 43.50±1.65 & 42.88±1.03 \\
Roman Empire    & 81.47±0.28 & 83.66±0.32          & \textbf{84.29±0.33} & 82.32±0.29 & 82.18±0.47 \\ \midrule[0.3pt]
\end{tabular}
}}
\end{table}

\begin{table}[t]
\setlength{\abovecaptionskip}{0.3cm}
\setlength{\belowcaptionskip}{0.2cm}
\caption{Ablation study on two node-wise attention mechanisms.
}
\label{tab: ab_exp}
\resizebox{\linewidth}{19mm}{
\setlength{\tabcolsep}{1mm}{
\begin{tabular}{ccccc}
\midrule[0.3pt]
Model             & CoraML              & CiteSeer            & Chameleon           & Squirrel            \\ \midrule[0.3pt]
w/o DP Attention  & 81.36±0.55          & 62.75±0.28          & 44.32±1.83          & 43.57±0.84          \\
ADPA-DP-Original    & \textbf{84.52±0.64} & \textbf{66.03±0.38} & 45.81±1.22          & 44.76±0.93          \\
ADPA-DP-Gate      & {82.37±0.78} & {65.27±0.32} & 45.87±1.46          & 44.96±1.12          \\
ADPA-DP-Recursive & 83.40±0.80          & 64.86±0.47          & 46.06±1.30          & \textbf{45.22±1.28} \\
ADPA-DP-JK     & 83.86±0.55          & 65.14±0.30          & \textbf{46.19±1.34} & {45.03±1.16} \\
w/o Hop Attention & 80.84±0.42          & 63.18±0.26          & {43.80±1.59} & 43.36±0.79          \\ \midrule[0.3pt]
ADPA              & 84.52±0.64          & 66.03±0.38          & 46.19±1.34          & 45.22±1.28          \\ \midrule[0.3pt]
\end{tabular}
}}
\end{table}

\subsection{Ablation Study and Sensitivity Analysis}
\label{sec: Ablation Study and Sensitivity Analysis}

    {To answer \textbf{Q3}, we focus on two critical modules introduced in our proposed ADPA: 
    (1) DP guided feature propagation; and 
    (2) two hierarchical node-wise attention mechanisms.
    The motivation and technical details of defining DP operators and utilizing them for feature propagation to encode deep structural information can be found in Sec.~\ref{sec: AMUD: Graph-based Data Engineering} and Sec.~\ref{sec: directed patterns guided feature propagation}. 
    }
    Notably, in the aforementioned process, we employ $k$-order DP operators at each propagation step to capture multi-scale directed topology (Level 1), thereby obtaining multi-granularity structural insights after completing the $K$-step propagation (Level 2).
    Building upon this foundation, we propose two hierarchical node-wise attention mechanisms: DP attention for Level 1 and Hop attention for Level 2. 
    The relevant technical details can be found in Sec.~\ref{sec: Node-adaptive Attention Mechanisms}. 
    The purpose of this strategy is to achieve efficient node representation fusion through an end-to-end learnable mechanism, improving predictive performance.

\noindent
\textbf{$k$-order DPs and $K$-step feature propagation.} 
    As depicted in Table~\ref{tab: ablation_k-orders}, we present the impact of various $k$-order DP operators on ADPA. 
    Notably, ADPA with $2$-order DP operators attains the optimal performance across most datasets, including CoraML, CiteSeer, Chameleon, Squirrel, and others. 
    Although $3$-order DP operators enhance performance on specific datasets, such as Actor and Amazon Rating, the utilization of higher-order DP operators does not yield any positive impact, instead, it adversely affects the model's performance.
    We attribute this phenomenon to over-fitting issues, which means that higher-order DP operators not only produce redundant structural encoding but also require more learnable weights for accurate prediction.
    Additionally, we observe that ADPA with $1$-order DP operators exhibits weak performance across the most of datasets. 
    This can be attributed to the limited expressive power of $1$-order DP operators, which only encompass 1-hop in- or out-neighbors of a node and consequently provide less information compared to $2$-order or higher DP operators.

    For an in-depth analysis, we illustrate the impact of various propagation steps on models (i.e., SGC, GPRGNN, NSTE, DIMPA, and our proposed ADPA) across AMUndirected datasets (CoraML, CiteSeer, and Actor) and AMDirected datasets (Cornell, Chameleon, and Squirrel) in Fig.~\ref{fig: exp_prop}. 
    We can conclude that most GNNs exhibit improved performance as the propagation step $K$ increases from 1 to 3. 
    However, after reaching the boundary point, introducing more propagation steps has a detrimental effect on the performance of most models, primarily due to the over-smoothing issues.
    To tackle this unique challenge, our proposed node-wise hop attention proves effective by allowing each node to adaptively aggregate propagated features according to its specific requirements.
    Consequently, ADPA consistently outperforms other models, even with an increased number of propagation steps.

\noindent
\textbf{Node-wise DP and hop attention.}
    {The ablation experiment for two hierarchical node-wise attention mechanisms is depicted in Table~\ref{tab: ab_exp}, where ADPA-DP-Original represents the attention mechanism employed in Eq.~(\ref{eq: node-wise dp attention}), and Gate, Recursive, and JK correspond to the attention mechanisms proposed in~\cite{ahmad2021gate},~\cite{gamlp}, and~\cite{xu2018jknet}, respectively.
    Notably, node-wise DP attention plays a crucial role in the predictive performance improvement of ADPA, contributing to an average accuracy gain of over 2\%.}
    As discussed in Sec.~\ref{sec: directed modeling guidance}, as different DP operators provide distinct structural insights for various semantic information, selectively aggregating features propagated by these operators is essential.
    Furthermore, Table~\ref{tab: ab_exp} reveals that the original attention mechanism used in Eq.~(\ref{eq: node-wise dp attention}) achieves optimal performance in AMUndirected datasets (CoraML and CiteSeer), while Recursive and JK attention attain the best performance in Chameleon and Squirrel of AMDirected datasets. 
    This discrepancy can be attributed to the homophily and heterophily characteristics of these datasets. 
    For instance, AMDirected datasets demand more learnable parameters attained in Recursive and JK attention to capture their intricate topological relationships.
    
    Meanwhile, node-wise hop attention also significantly contributes to the performance improvement, yielding an average gain of over 3\%. 
    In contrast to JK attention, our proposed node-wise hop attention in Eq.~(\ref{eq: JK attention}) aggregates features from different DP operators separately. 
    As a result, it possesses a more semantically diverse and multi-scale reception field than JK attention, leading to superior performance in predictions.

\begin{figure}[t]
	\centering
    \setlength{\abovecaptionskip}{0.2cm}
    \setlength{\belowcaptionskip}{-0.2cm}
  \includegraphics[width=\linewidth]{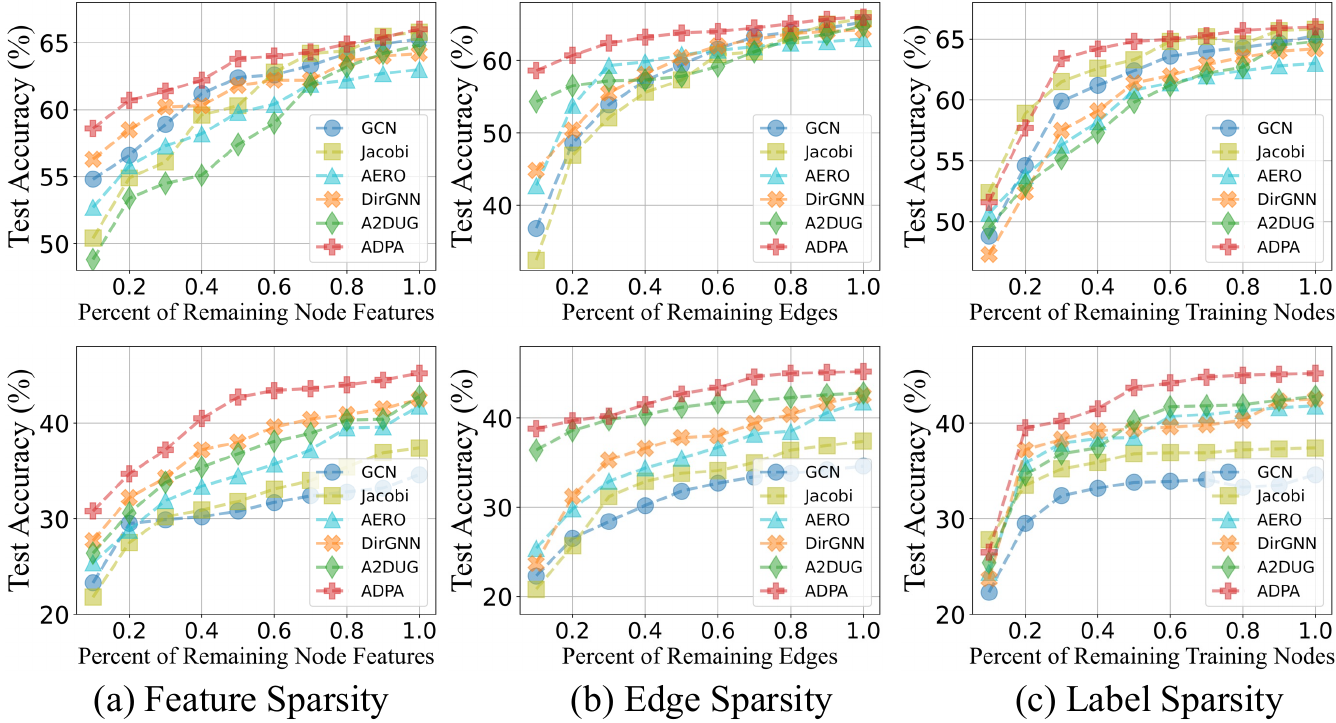}
  \caption{
  Performance on CiteSeer (upper) and Squirrel (lower) under different levels of sparsity.}

  \label{fig:exp_sparsity}
\end{figure}
\subsection{Performance on Sparse Settings}
\label{Performance on Sparse Settings}
    To answer \textbf{Q4}, we provide experimental results in Fig.~\ref{fig:exp_sparsity}.
    Given the practical challenges in the real world, a notable concern is the inherent sparsity, particularly in digraphs with abundant directed information from intricate applications. 
    This is because 
    (1) compared to benchmark datasets used in academic research, we struggle to construct high-quality node features in the industry; 
    (2) the precise description of topology by directed edges reduces redundant structural information;
    (3) in the semi-supervised node classification paradigm, where only a small proportion of nodes are labeled, label sparsity becomes a realistic issue.
    The aforementioned features, edge, and label sparsity pose unique challenges for existing DiGNNs.
    
    For feature sparsity, we assume that the feature representation of unlabeled nodes is partially missing. 
    In this case, it is necessary to obtain additional feature information from neighbors through appropriate propagation. 
    Fig.~\ref{fig:exp_sparsity} shows that A2DUG may suffer from limited RF due to the lack of utilizing digraph structure for feature propagation, which leads to sub-optimal performance. 
    Notably, this similar approach includes LINK, which is not presented in Fig.~\ref{fig:exp_sparsity} to avoid complex charts and make results more reader-friendly.
    These methods directly fed digraph topology to $\operatorname{MLP}$ to encode structural information and combine it with node feature embeddings to avoid exploring intricate relationships in digraphs. 
    While providing a straightforward and efficient solution, these methods often suffer from degraded predictive performance because they overlook explicit interactions between nodes.
    Additionally, they have a notable drawback—they cannot address the issue of feature sparsity.
    However, they exhibit unexpected benefits in the context of edge sparsity.
    Moreover, methods exemplified by JacobiConv heavily rely on node features to conduct spectral analysis for discovering connection patterns between node pairs. 
    This dependence results in unacceptable performance degradation in sparse feature settings.
    On the contrary, ADPA and DirGNN can customize the number of propagation steps to achieve a larger RF, thus alleviating the problem of feature sparsity, which is also applicable to edge and label-sparse scenarios.
    To simulate edge sparsity, we randomly remove a fixed percentage of edges from the original digraph, providing a realistic challenge. 
    For label sparsity, we change the number of labeled samples for each class. 
    Experimental results from Fig.~\ref{fig:exp_sparsity} show that our proposed ADPA, as compared to baselines, is more robust to the sparsity scenarios.

\vspace{0.1cm}

\section{Conclusion}
    In this paper, we firstly review recent developments in graph learning based on the semi-supervised node classification paradigm. 
    Inspired by practical demands of realistic applications, we emphasize that the intricate topology of graphs database places existing GNNs in a dilemma in exploring the optimal node representation due to the entanglement of homophily and heterophily. 
    To solve this issue, we propose AMUD to provide a new perspective of data engineering by utilizing the directed information, modeling the intricate topology to break such an entanglement.  
    Specifically, AMUD quantifies the correlation between node profiles and topology and guides whether to retain the inherent directed edges to maximize the benefits of subsequent graph learning as shown in Fig.~\ref{fig: motivation_AMUD_AdpA}. 
    Additionally, we propose a new digraph learning paradigm ADPA, which explores the optimal node representation in directed scenarios by adaptively mining the suitable DPs and employing two hierarchical node-wise attention mechanisms. ADPA exhibits statisfying results in both digraphs and undirected graphs. 
    Promising future directions are to explore a more effective propagation mechanism that can achieve fine-grained exploration of complex topology and a more efficient execution of multi-scale information fusion while maintaining performance.



\newpage
\balance{
\bibliographystyle{IEEEtran}
\bibliography{IEEEexample}

\begin{thebibliography}{10}
\providecommand{\url}[1]{#1}
\csname url@samestyle\endcsname
\providecommand{\newblock}{\relax}
\providecommand{\bibinfo}[2]{#2}
\providecommand{\BIBentrySTDinterwordspacing}{\spaceskip=0pt\relax}
\providecommand{\BIBentryALTinterwordstretchfactor}{4}
\providecommand{\BIBentryALTinterwordspacing}{\spaceskip=\fontdimen2\font plus
\BIBentryALTinterwordstretchfactor\fontdimen3\font minus \fontdimen4\font\relax}
\providecommand{\BIBforeignlanguage}[2]{{%
\expandafter\ifx\csname l@#1\endcsname\relax
\typeout{** WARNING: IEEEtran.bst: No hyphenation pattern has been}%
\typeout{** loaded for the language `#1'. Using the pattern for}%
\typeout{** the default language instead.}%
\else
\language=\csname l@#1\endcsname
\fi
#2}}
\providecommand{\BIBdecl}{\relax}
\BIBdecl

\bibitem{guan2023homo_measure_icde2}
S.~Guan, H.~Ma, M.~Wang, and Y.~Wu, ``Gale: Active adversarial learning for erroneous node detection in graphs,'' in \emph{IEEE 39th International Conference on Data Engineering, ICDE}.\hskip 1em plus 0.5em minus 0.4em\relax IEEE, 2023.

\bibitem{li2023irreversible_graph_vldb2}
Y.~Li, Y.~Shen, L.~Chen, and M.~Yuan, ``Zebra: When temporal graph neural networks meet temporal personalized pagerank,'' \emph{Proceedings of the VLDB Endowment}, 2023.

\bibitem{he2023scaling_sigmod2}
Y.~He, K.~Wang, W.~Zhang, X.~Lin, and Y.~Zhang, ``Scaling up k-clique densest subgraph detection,'' \emph{Proceedings of the ACM on Management of Data}, 2023.

\bibitem{10.1007/s00521-022-07735-y_EHGCN}
X.~Li, R.~Guo, J.~Chen, Y.~Hu, M.~Qu, and B.~Jiang, ``Effective hybrid graph and hypergraph convolution network for collaborative filtering,'' \emph{Neural Comput. Appl.}, vol.~35, no.~3, p. 2633–2646, sep 2022.

\bibitem{xia2023app_gnn_rec1}
L.~Xia, C.~Huang, J.~Shi, and Y.~Xu, ``Graph-less collaborative filtering,'' in \emph{Proceedings of the ACM Web Conference, WWW}, 2023, pp. 17--27.

\bibitem{yang2023app_gnn_rec2}
L.~Yang, S.~Wang, Y.~Tao, J.~Sun, X.~Liu, P.~S. Yu, and T.~Wang, ``Dgrec: Graph neural network for recommendation with diversified embedding generation,'' in \emph{Proceedings of the Sixteenth ACM International Conference on Web Search and Data Mining, WSDM}, 2023, pp. 661--669.

\bibitem{bang2023app_gnn_bio1}
D.~Bang, S.~Lim, S.~Lee, and S.~Kim, ``Biomedical knowledge graph learning for drug repurposing by extending guilt-by-association to multiple layers,'' \emph{Nature Communications}, vol.~14, no.~1, p. 3570, 2023.

\bibitem{qu2023app_gnn_bio2}
Z.~Qu, T.~Yao, X.~Liu, and G.~Wang, ``A graph convolutional network based on univariate neurodegeneration biomarker for alzheimer’s disease diagnosis,'' \emph{IEEE Journal of Translational Engineering in Health and Medicine}, 2023.

\bibitem{gao2023app_gnn_bio3}
Z.~Gao, H.~Ma, X.~Zhang, Y.~Wang, and Z.~Wu, ``Similarity measures-based graph co-contrastive learning for drug--disease association prediction,'' \emph{Bioinformatics}, vol.~39, no.~6, p. btad357, 2023.

\bibitem{tang2022app_detection1}
J.~Tang, J.~Li, Z.~Gao, and J.~Li, ``Rethinking graph neural networks for anomaly detection,'' in \emph{International Conference on Machine Learning, ICML}.\hskip 1em plus 0.5em minus 0.4em\relax PMLR, 2022, pp. 21\,076--21\,089.

\bibitem{chen2022app_detection2}
B.~Chen, J.~Zhang, X.~Zhang, Y.~Dong, J.~Song, P.~Zhang, K.~Xu, E.~Kharlamov, and J.~Tang, ``Gccad: Graph contrastive learning for anomaly detection,'' \emph{IEEE Transactions on Knowledge and Data Engineering}, 2022.

\bibitem{duan2023app_detection3}
J.~Duan, S.~Wang, P.~Zhang, E.~Zhu, J.~Hu, H.~Jin, Y.~Liu, and Z.~Dong, ``Graph anomaly detection via multi-scale contrastive learning networks with augmented view,'' in \emph{Proceedings of the AAAI Conference on Artificial Intelligence, AAAI}, vol.~37, no.~6, 2023, pp. 7459--7467.

\bibitem{wang2020gcnlpa}
H.~Wang and J.~Leskovec, ``Unifying graph convolutional neural networks and label propagation,'' \emph{arXiv preprint arXiv:2002.06755}, 2020.

\bibitem{chen2020gcnii}
M.~Chen, Z.~Wei, Z.~Huang, B.~Ding, and Y.~Li, ``Simple and deep graph convolutional networks,'' in \emph{International Conference on Machine Learning, ICML}, 2020.

\bibitem{gamlp}
W.~Zhang, Z.~Yin, Z.~Sheng, Y.~Li, W.~Ouyang, X.~Li, Y.~Tao, Z.~Yang, and B.~Cui, ``Graph attention multi-layer perceptron,'' \emph{Proceedings of the 28th ACM SIGKDD Conference on Knowledge Discovery and Data Mining, KDD}, 2022.

\bibitem{ye2023homo_measure_icde1}
J.~Ye, Y.~Zhu, and L.~Chen, ``Top-r keyword-based community search in attributed graphs,'' in \emph{IEEE 39th International Conference on Data Engineering, ICDE}.\hskip 1em plus 0.5em minus 0.4em\relax IEEE, 2023.

\bibitem{wu2023billionambiguous_graph_de_vldb1}
X.~Wu, Y.~Xu, W.~Zhang, and Y.~Zhang, ``Billion-scale bipartite graph embedding: A global-local induced approach,'' \emph{Proceedings of the VLDB Endowment}, 2023.

\bibitem{liu2023efficientintro_sigmod1}
K.~Liu, S.~Wang, Y.~Zhang, and C.~Xing, ``An efficient algorithm for distance-based structural graph clustering,'' \emph{Proceedings of the ACM on Management of Data}, vol.~1, no.~1, pp. 1--25, 2023.

\bibitem{guan2023homo_measure_icde3}
G.~Zhu, Z.~Zhu, W.~Wang, Z.~Xu, C.~Yuan, and Y.~Huang, ``Autoac: Towards automated attribute completion for heterogeneous graph neural network,'' in \emph{IEEE 39th International Conference on Data Engineering, ICDE}.\hskip 1em plus 0.5em minus 0.4em\relax IEEE, 2023.

\bibitem{yin2023surel+irreversible_graph_vldb3}
H.~Yin, M.~Zhang, J.~Wang, and P.~Li, ``Surel+: Moving from walks to sets for scalable subgraph-based graph representation learning,'' \emph{Proceedings of the VLDB Endowment}, 2023.

\bibitem{li2023efficient_sigmod3}
Y.~Li, R.~Yang, and J.~Shi, ``Efficient and effective attributed hypergraph clustering via k-nearest neighbor augmentation,'' \emph{Proceedings of the ACM on Management of Data}, 2023.

\bibitem{mcpherson200homophily_theory1}
M.~McPherson, L.~Smith-Lovin, and J.~M. Cook, ``Birds of a feather: Homophily in social networks,'' \emph{Annual review of socioLoGy}, vol.~27, no.~1, pp. 415--444, 2001.

\bibitem{M2003Mixing_homophily_theory2}
M., E., J., and Newman, ``Mixing patterns in networks,'' \emph{Physical Review E}, vol.~67, no.~2, pp. 26\,126--26\,126, 2003.

\bibitem{0Networks_homophily_theory3}
O.~C. E. O.~B. Author@Ac, \emph{Networks, Crowds, and Markets}.\hskip 1em plus 0.5em minus 0.4em\relax Networks, Crowds, and Markets.

\bibitem{wu2020gnn_survey1}
Z.~Wu, S.~Pan, F.~Chen, G.~Long, C.~Zhang, and S.~Y. Philip, ``A comprehensive survey on graph neural networks,'' \emph{IEEE transactions on neural networks and learning systems}, vol.~32, no.~1, pp. 4--24, 2020.

\bibitem{zhou2022gnn_survey2}
Y.~Zhou, H.~Zheng, X.~Huang, S.~Hao, D.~Li, and J.~Zhao, ``Graph neural networks: Taxonomy, advances, and trends,'' \emph{ACM Transactions on Intelligent Systems and TechnoLoGy (TIST)}, vol.~13, no.~1, pp. 1--54, 2022.

\bibitem{bessadok2022gnn_survey3}
A.~Bessadok, M.~A. Mahjoub, and I.~Rekik, ``Graph neural networks in network neuroscience,'' \emph{IEEE Transactions on Pattern Analysis and Machine Intelligence}, vol.~45, no.~5, pp. 5833--5848, 2022.

\bibitem{ma2021hete_gnn_survey1}
Y.~Ma, X.~Liu, N.~Shah, and J.~Tang, ``Is homophily a necessity for graph neural networks?'' \emph{International Conference on Learning Representations, ICLR}, 2021.

\bibitem{luan2022hete_gnn_survey2}
S.~Luan, C.~Hua, Q.~Lu, J.~Zhu, M.~Zhao, S.~Zhang, X.-W. Chang, and D.~Precup, ``Revisiting heterophily for graph neural networks,'' \emph{Advances in neural information processing systems, NeurIPS}, 2022.

\bibitem{zheng2022hete_gnn_survey3}
X.~Zheng, Y.~Liu, S.~Pan, M.~Zhang, D.~Jin, and P.~S. Yu, ``Graph neural networks for graphs with heterophily: A survey,'' \emph{arXiv preprint arXiv:2202.07082}, 2022.

\bibitem{he2021bernnet}
M.~He, Z.~Wei, H.~Xu \emph{et~al.}, ``Bernnet: Learning arbitrary graph spectral filters via bernstein approximation,'' \emph{Advances in Neural Information Processing Systems, NeurIPS}, 2021.

\bibitem{pmlr2022Jacobigcn}
X.~Wang and M.~Zhang, ``How powerful are spectral graph neural networks,'' in \emph{Proceedings of the 39th International Conference on Machine Learning, ICML}, 2022.

\bibitem{guo2023optimal_poly_gnn}
Y.~Guo and Z.~Wei, ``Graph neural networks with learnable and optimal polynomial bases,'' \emph{International Conference on Machine Learning, ICML}, 2023.

\bibitem{2022glognn}
X.~Li, R.~Zhu, Y.~Cheng, C.~Shan, S.~Luo, D.~Li, and W.~Qian, ``Finding global homophily in graph neural networks when meeting heterophily,'' 2022.

\bibitem{lee2023aerognn}
S.~Y. Lee, F.~Bu, J.~Yoo, and K.~Shin, ``Towards deep attention in graph neural networks: Problems and remedies,'' in \emph{Proceedings of the 40th International Conference on Machine Learning, ICML}, 2023.

\bibitem{yoo2023slimg}
J.~Yoo, M.-C. Lee, S.~Shekhar, and C.~Faloutsos, ``Less is more: Slimg for accurate, robust, and interpretable graph mining,'' in \emph{Proceedings of the 29th ACM SIGKDD Conference on Knowledge Discovery and Data Mining, KDD}, 2023.

\bibitem{dirgnn_rossi_2023}
E.~Rossi, B.~Charpentier, F.~D. Giovanni, F.~Frasca, S.~Günnemann, and M.~Bronstein, ``Edge directionality improves learning on heterophilic graphs,'' \emph{in Proceedings of The European Conference on Machine Learning and Principles and Practice of Knowledge Discovery in Databases , ECML-PKDD Workshop}, 2023.

\bibitem{maekawa2023a2dug}
S.~Maekawa, Y.~Sasaki, and M.~Onizuka, ``Why using either aggregated features or adjacency lists in directed or undirected graph? empirical study and simple classification method,'' \emph{arXiv preprint arXiv:2306.08274}, 2023.

\bibitem{chung2005spectral_graph_magnetic_laplacian1}
F.~Chung, ``Laplacians and the cheeger inequality for directed graphs,'' \emph{Annals of Combinatorics}, vol.~9, pp. 1--19, 2005.

\bibitem{kipf2016gcn}
T.~N. Kipf and M.~Welling, ``Semi-supervised classification with graph convolutional networks,'' in \emph{International Conference on Learning Representations, ICLR}, 2017.

\bibitem{10.5555/2968618.2968693_homophily_theory4}
O.~Chapelle, J.~Weston, and B.~Sch\"{o}lkopf, ``Cluster kernels for semi-supervised learning,'' in \emph{Proceedings of the 15th International Conference on Neural Information Processing Systems, NeurIPS}.\hskip 1em plus 0.5em minus 0.4em\relax Cambridge, MA, USA: MIT Press, 2002, p. 601–608.

\bibitem{xu2018jknet}
K.~Xu, C.~Li, Y.~Tian, T.~Sonobe, K.-i. Kawarabayashi, and S.~Jegelka, ``Representation learning on graphs with jumping knowledge networks,'' in \emph{International conference on machine learning}.\hskip 1em plus 0.5em minus 0.4em\relax PMLR, 2018, pp. 5453--5462.

\bibitem{zeng2019graphsaint}
H.~Zeng, H.~Zhou, A.~Srivastava, R.~Kannan, and V.~Prasanna, ``Graphsaint: Graph sampling based inductive learning method,'' in \emph{International conference on learning representations, ICLR}, 2020.

\bibitem{hamilton2017graphsage}
W.~Hamilton, Z.~Ying, and J.~Leskovec, ``Inductive representation learning on large graphs,'' \emph{Advances in Neural Information Processing Systems, NeurIPS}, 2017.

\bibitem{2019appnp}
J.~Klicpera, A.~Bojchevski, and S.~Günnemann, ``Predict then propagate: Graph neural networks meet personalized pagerank,'' in \emph{International Conference on Learning Representations, ICLR}, 2019.

\bibitem{Hu2021ahgae}
Y.~Hu, X.~Li, Y.~Wang, Y.~Wu, Y.~Zhao, C.~Yan, J.~Yin, and Y.~Gao, ``Adaptive hypergraph auto-encoder for relational data clustering,'' \emph{IEEE Transactions on Knowledge and Data Engineering}, 2021.

\bibitem{luo2022ambiguous_graph_de_icde5}
X.~Luo, W.~Ju, M.~Qu, C.~Chen, M.~Deng, X.-S. Hua, and M.~Zhang, ``Dualgraph: Improving semi-supervised graph classification via dual contrastive learning,'' in \emph{IEEE 38th International Conference on Data Engineering, ICDE}.\hskip 1em plus 0.5em minus 0.4em\relax IEEE, 2022.

\bibitem{lv2023datairreversible_graph_vldb5}
G.~Lv and L.~Chen, ``On data-aware global explainability of graph neural networks,'' \emph{Proceedings of the VLDB Endowment}, 2023.

\bibitem{wang2023scapin_sigmod5}
Y.~Wang, Z.~Yang, J.~Liu, W.~Zhang, and B.~Cui, ``Scapin: Scalable graph structure perturbation by augmented influence maximization,'' \emph{Proceedings of the ACM on Management of Data}, 2023.

\bibitem{chien2021gprgnn}
E.~Chien, J.~Peng, P.~Li, and O.~Milenkovic, ``Adaptive universal generalized pagerank graph neural network,'' in \emph{International Conference on Learning Representations, ICLR}, 2021.

\bibitem{2021linkx}
D.~Lim, F.~Hohne, X.~Li, S.~L. Huang, V.~Gupta, O.~Bhalerao, and S.~N. Lim, ``Large scale learning on non-homophilous graphs: New benchmarks and strong simple methods,'' in \emph{arXiv}, 2021.

\bibitem{pei2020geomgcn}
H.~Pei, B.~Wei, K.~C.-C. Chang, Y.~Lei, and B.~Yang, ``Geom-gcn: Geometric graph convolutional networks,'' in \emph{International Conference on Learning Representations, ICLR}, 2020.

\bibitem{yan2021ggcn}
Y.~Yan, M.~Hashemi, K.~Swersky, Y.~Yang, and D.~Koutra, ``Two sides of the same coin: Heterophily and oversmoothing in graph convolutional neural networks,'' \emph{arXiv preprint arXiv:2102.06462}, 2022.

\bibitem{dai2022lwgcn}
E.~Dai, S.~Zhou, Z.~Guo, and S.~Wang, ``Label-wise graph convolutional network for heterophilic graphs,'' in \emph{Learning on Graphs Conference, LoG}.\hskip 1em plus 0.5em minus 0.4em\relax PMLR, 2022, pp. 26--1.

\bibitem{du2022gbkgnn}
L.~Du, X.~Shi, Q.~Fu, X.~Ma, H.~Liu, S.~Han, and D.~Zhang, ``Gbk-gnn: Gated bi-kernel graph neural networks for modeling both homophily and heterophily,'' in \emph{Proceedings of the ACM Web Conference, WWW}, 2022, pp. 1550--1558.

\bibitem{song2023ordergnn}
Y.~Song, C.~Zhou, X.~Wang, and Z.~Lin, ``Ordered gnn: Ordering message passing to deal with heterophily and over-smoothing,'' \emph{International conference on learning representations, ICLR}, 2023.

\bibitem{zeng2022ambiguous_graph_de_icde4}
J.~Zeng, P.~Wang, L.~Lan, J.~Zhao, F.~Sun, J.~Tao, J.~Feng, M.~Hu, and X.~Guan, ``Accurate and scalable graph neural networks for billion-scale graphs,'' in \emph{IEEE 38th International Conference on Data Engineering, ICDE}.\hskip 1em plus 0.5em minus 0.4em\relax IEEE, 2022.

\bibitem{lv2023henceirreversible_graph_vldb4}
G.~Lv, C.~J. Zhang, and L.~Chen, ``Hence-x: Toward heterogeneity-agnostic multi-level explainability for deep graph networks,'' \emph{Proceedings of the VLDB Endowment}, 2023.

\bibitem{wan2023scalable_sigmod4}
X.~Wan, K.~Xu, X.~Liao, Y.~Jin, K.~Chen, and X.~Jin, ``Scalable and efficient full-graph gnn training for large graphs,'' \emph{Proceedings of the ACM on Management of Data}, 2023.

\bibitem{2020h2gcn}
J.~Zhu, Y.~Yan, L.~Zhao, M.~Heimann, L.~Akoglu, and D.~Koutra, ``Beyond homophily in graph neural networks: Current limitations and effective designs,'' \emph{Advances in Neural Information Processing Systems, NeurIPS}, 2020.

\bibitem{platonov2023hete_gnn_survey4}
O.~Platonov, D.~Kuznedelev, M.~Diskin, A.~Babenko, and L.~Prokhorenkova, ``A critical look at the evaluation of gnns under heterophily: are we really making progress?'' \emph{International Conference on Learning Representations, ICLR}, 2023.

\bibitem{platonov2022hete_gnn_survey5}
O.~Platonov, D.~Kuznedelev, A.~Babenko, and L.~Prokhorenkova, ``Characterizing graph datasets for node classification: Beyond homophily-heterophily dichotomy,'' \emph{Advances in Neural Information Processing Systems, NeurIPS}, 2023.

\bibitem{tong2020dgcn}
Z.~Tong, Y.~Liang, C.~Sun, D.~S. Rosenblum, and A.~Lim, ``Directed graph convolutional network,'' \emph{arXiv preprint arXiv:2004.13970}, 2020.

\bibitem{he2022dimpa}
Y.~He, G.~Reinert, and M.~Cucuringu, ``Digrac: Digraph clustering based on flow imbalance,'' in \emph{Learning on Graphs Conference, LoG}.\hskip 1em plus 0.5em minus 0.4em\relax PMLR, 2022, pp. 21--1.

\bibitem{kollias2022nste}
G.~Kollias, V.~Kalantzis, T.~Id{\'e}, A.~Lozano, and N.~Abe, ``Directed graph auto-encoders,'' in \emph{Proceedings of the Conference on Artificial Intelligence, AAAI}, vol.~36, no.~7, 2022, pp. 7211--7219.

\bibitem{tong2020digcn}
Z.~Tong, Y.~Liang, C.~Sun, X.~Li, D.~Rosenblum, and A.~Lim, ``Digraph inception convolutional networks,'' \emph{Advances in neural information processing systems, NeurIPS}, vol.~33, pp. 17\,907--17\,918, 2020.

\bibitem{zhang2021magnet}
X.~Zhang, Y.~He, N.~Brugnone, M.~Perlmutter, and M.~Hirn, ``Magnet: A neural network for directed graphs,'' \emph{Advances in neural information processing systems, NeurIPS}, vol.~34, pp. 27\,003--27\,015, 2021.

\bibitem{zhang2021mgc}
J.~Zhang, B.~Hui, P.-W. Harn, M.-T. Sun, and W.-S. Ku, ``Mgc: A complex-valued graph convolutional network for directed graphs,'' \emph{arXiv e-prints}, pp. arXiv--2110, 2021.

\bibitem{baeza2006linear_rank}
R.~Baeza-Yates, P.~Boldi, and C.~Castillo, ``Generalizing pagerank: Damping functions for link-based ranking algorithms,'' in \emph{Proceedings of the 29th annual international ACM SIGIR conference on Research and development in information retrieval, SIGIR}, 2006, pp. 308--315.

\bibitem{song2022gnn_survey4}
Z.~Song, X.~Yang, Z.~Xu, and I.~King, ``Graph-based semi-supervised learning: A comprehensive review,'' \emph{IEEE Transactions on Neural Networks and Learning Systems}, 2022.

\bibitem{milo2002_network_motif_1}
R.~Milo, S.~Shen-Orr, S.~Itzkovitz, N.~Kashtan, D.~Chklovskii, and U.~Alon, ``Network motifs: simple building blocks of complex networks,'' \emph{Science}, vol. 298, no. 5594, pp. 824--827, 2002.

\bibitem{benson2016_network_motif_2}
A.~R. Benson, D.~F. Gleich, and J.~Leskovec, ``Higher-order organization of complex networks,'' \emph{Science}, vol. 353, no. 6295, pp. 163--166, 2016.

\bibitem{ribeiro2021_network_motif_3}
P.~Ribeiro, P.~Paredes, M.~E. Silva, D.~Aparicio, and F.~Silva, ``A survey on subgraph counting: concepts, algorithms, and applications to network motifs and graphlets,'' \emph{ACM Computing Surveys, CSUR}, vol.~54, no.~2, pp. 1--36, 2021.

\bibitem{wu2019sgc}
F.~Wu, A.~Souza, T.~Zhang, C.~Fifty, T.~Yu, and K.~Weinberger, ``Simplifying graph convolutional networks,'' in \emph{International conference on machine learning, ICML}, 2019.

\bibitem{frasca2020sign}
F.~Frasca, E.~Rossi, D.~Eynard, B.~Chamberlain, M.~Bronstein, and F.~Monti, ``Sign: Scalable inception graph neural networks,'' \emph{arXiv preprint arXiv:2004.11198}, 2020.

\bibitem{sun2021sagn}
C.~Sun, H.~Gu, and J.~Hu, ``Scalable and adaptive graph neural networks with self-label-enhanced training,'' \emph{arXiv preprint arXiv:2104.09376}, 2021.

\bibitem{zhu2021ssgc}
H.~Zhu and P.~Koniusz, ``Simple spectral graph convolution,'' in \emph{International conference on learning representations, ICLR}, 2021.

\bibitem{zhang2021rod}
W.~Zhang, Y.~Jiang, Y.~Li, Z.~Sheng, Y.~Shen, X.~Miao, L.~Wang, Z.~Yang, and B.~Cui, ``Rod: reception-aware online distillation for sparse graphs,'' in \emph{Proceedings of the 27th ACM SIGKDD Conference on Knowledge Discovery \& Data Mining, KDD}, 2021.

\bibitem{zhang2022air}
W.~Zhang, Z.~Sheng, Z.~Yin, Y.~Jiang, Y.~Xia, J.~Gao, Z.~Yang, and B.~Cui, ``Model degradation hinders deep graph neural networks,'' in \emph{Proceedings of the 28th ACM SIGKDD Conference on Knowledge Discovery and Data Mining, KDD}, 2022.

\bibitem{ahmad2021gate}
W.~U. Ahmad, N.~Peng, and K.-W. Chang, ``Gate: graph attention transformer encoder for cross-lingual relation and event extraction,'' in \emph{Proceedings of the AAAI Conference on Artificial Intelligence}, vol.~35, no.~14, 2021, pp. 12\,462--12\,470.

\bibitem{2021fsgnn}
S.~K. Maurya, X.~Liu, and T.~Murata, ``Improving graph neural networks with simple architecture design,'' 2021.

\bibitem{bojchevski2018coraml_citeseer}
\BIBentryALTinterwordspacing
A.~Bojchevski and S.~Günnemann, ``Deep gaussian embedding of graphs: Unsupervised inductive learning via ranking,'' in \emph{ICLR Workshop on Representation Learning on Graphs and Manifolds}, 2018. [Online]. Available: \url{https://openreview.net/forum?id=r1ZdKJ-0W}
\BIBentrySTDinterwordspacing

\bibitem{Yang16cora}
Z.~Yang, W.~W. Cohen, and R.~Salakhutdinov, ``Revisiting semi-supervised learning with graph embeddings,'' in \emph{Proceedings of the 33rd International Conference on International Conference on Machine Learning, ICML}, 2016, p. 40–48.

\bibitem{hu2020ogb}
W.~Hu, M.~Fey, M.~Zitnik, Y.~Dong, H.~Ren, B.~Liu, M.~Catasta, and J.~Leskovec, ``Open graph benchmark: Datasets for machine learning on graphs,'' \emph{Advances in neural information processing systems, NeurIPS}, vol.~33, pp. 22\,118--22\,133, 2020.

\bibitem{shchur2018amazon_datasets}
O.~Shchur, M.~Mumme, A.~Bojchevski, and S.~G{\"u}nnemann, ``Pitfalls of graph neural network evaluation,'' \emph{arXiv preprint arXiv:1811.05868}, 2018.

\bibitem{chamberlain2021grand}
B.~Chamberlain, J.~Rowbottom, M.~I. Gorinova, M.~Bronstein, S.~Webb, and E.~Rossi, ``Grand: Graph neural diffusion,'' in \emph{International Conference on Machine Learning, ICML}.\hskip 1em plus 0.5em minus 0.4em\relax PMLR, 2021.

\bibitem{akiba2019optuna}
T.~Akiba, S.~Sano, T.~Yanase, T.~Ohta, and M.~Koyama, ``Optuna: A next-generation hyperparameter optimization framework,'' in \emph{Proceedings of the 25th ACM SIGKDD international conference on knowledge discovery \& data mining, KDD}, 2019, pp. 2623--2631.

\end{thebibliography}
}

\end{document}